\documentclass[10pt]{article} 
\usepackage[preprint]{tmlr}

\usepackage{amsmath,amsfonts,bm}

\def\eqref#1{equation~\ref{#1}}

\def\1{\bm{1}}

\DeclareMathAlphabet{\mathsfit}{\encodingdefault}{\sfdefault}{m}{sl}
\SetMathAlphabet{\mathsfit}{bold}{\encodingdefault}{\sfdefault}{bx}{n}

\usepackage{hyperref}
\usepackage{url}

\usepackage{microtype}
\usepackage{graphicx}
\usepackage{subfigure}
\usepackage{booktabs} 
\usepackage[page,header]{appendix}
\usepackage{titletoc}
\usepackage{minitoc}

\usepackage{color}
\usepackage{amsmath}
\usepackage{amssymb}
\usepackage{mathtools}
\usepackage{amsthm}
\usepackage{bm}
\usepackage{multirow}
\usepackage{subfigure}
\usepackage{booktabs}
\usepackage{wrapfig}

\usepackage{algorithm}
\usepackage{algorithmic}
\usepackage{amsmath}
\usepackage{xcolor}

\usepackage{tikz}
\usetikzlibrary{fit, tikzmark, calc}

\newcommand\drawCodeBox[3]{%
    \begin{tikzpicture}[remember picture,overlay]
        \coordinate (start) at ([yshift=1.7ex]pic cs:#2);
        \coordinate (end) at ([yshift=-0.8ex]pic cs:#3);
        \node[inner sep=2pt,#1,fit=(start) (end),rectangle, rounded corners=1mm, fill opacity=0.2] {};
    \end{tikzpicture}%
}

\newtheorem{theorem}{Theorem}[section]

\newtheorem{lemma}{Lemma}[section]
\newtheorem{corollary}{Corollary}[section]
\newtheorem{assumption}{Assumption}[section]
\newtheorem{example}{Example}[section]
\newtheorem{definition}{Definition}[section]
\newtheorem{remark}{Remark}[section]

\newcommand{\bs}[1]{\boldsymbol{#1}}
\newcommand{\ie}{\textit{i.e.,}}

\usepackage{amsthm}
\newenvironment{proofsketch}{%
  \proof}{\endproof}

\title{Enhanced Federated Optimization: \\ Adaptive Unbiased Client Sampling with Reduced Variance}

\author{\name Dun Zeng \email zengdun@std.uestc.edu.cn \\
      \addr University of Electronic Science and Technology of China
      \ANDD
      \name Zenglin Xu \email zenglinxu@fudan.edu.cn \\
      \addr Fudan University, Shanghai Academy of AI for Science
      \ANDD
      \name Yu Pan \email iperryuu@gmail.com\\
      \addr Harbin Institute of Technology, Shenzhen
      \ANDD
      \name Xu Luo \email frank.luox@outlook.com\\
      \addr University of Electronic Science and Technology of China
      \ANDD
      \name Qifan Wang \email wqfcr618@gmail.com\\
      \addr Meta AI
      \ANDD
      \name Xiaoying Tang \email tangxiaoying@cuhk.edu.cn\\
      \addr The Chinese University of Hong Kong, Shenzhen
}

\def\month{MM}  
\def\year{YYYY} 
\def\openreview{\url{https://openreview.net/forum?id=XXXX}} 

\begin{document}

\maketitle

\begin{abstract}
Federated Learning (FL) is a distributed learning paradigm to train a global model across multiple devices without collecting local data.  In FL, a server typically selects a subset of clients for each training round to optimize resource usage. Central to this process is the technique of unbiased client sampling, which ensures a representative selection of clients. Current methods primarily utilize a random sampling procedure which, despite its effectiveness, achieves suboptimal efficiency owing to the loose upper bound caused by the sampling variance. In this work, by adopting an independent sampling procedure, we propose a federated optimization framework focused on adaptive unbiased client sampling, improving the convergence rate via an online variance reduction strategy.
In particular, we present the first adaptive client sampler, K-Vib, employing an independent sampling procedure. K-Vib achieves a linear speed-up on the regret bound $\tilde{\mathcal{O}}\big(N^{\frac{1}{3}}T^{\frac{2}{3}}/K^{\frac{4}{3}}\big)$ within a set communication budget $K$. Empirical studies indicate that K-Vib doubles the speed compared to baseline algorithms, demonstrating significant potential in federated optimization.
\end{abstract}

\section{Introduction}
This paper studies  the prevalent cross-device federated learning (FL) framework, as outlined in~\citet{kairouz2021advances}, which optimizes $\bs{x} \in \mathcal{X} \subseteq \mathbb{R}^d$ to minimize a finite-sum objective:
\begin{equation}\label{eq:obj}
    \min_{\bs{x} \in \mathcal{X}} f(\bs{x}) := \sum_{i=1}^N\bs{\lambda}_i f_i(\bs{x}) := \sum_{i=1}^N\bs{\lambda}_i \mathbb{E}_{\xi_i \sim \mathcal{D}_i}[F_i(\bs{x}, \xi_i)],
\end{equation}
where $N$ denotes  the total number of clients, and $\bs{\lambda}$ denotes the weights of client objective  ($\bs{\lambda}_i\geq 0, \sum_{i=1}^N \bs{\lambda}_i =1 $ ). The local loss function $f_i:\mathbb{R}^d \rightarrow \mathbb{R}$  is intricately linked to the local data distribution $D_i$. It is defined as $f_i(\bs{x}) = \mathbb{E}_{\xi_i \sim \mathcal{D}_i}[F_i(\bs{x}, \xi_i)]$, where $\xi_i$ 
represents a stochastic batch drawn from $D_i$. Federated optimization algorithms, such as \textsc{FedAvg}~\citep{mcmahan2017communication}, are designed to minimize objectives like \eqref{eq:obj} by alternating between local and global updates in a distributed learning framework. To reduce communication and computational demands in FL~\citep{konevcny2016federated, wang2021field, yang2022practical}, various client sampling strategies have been developed~\citep{chen2020optimal, cho2020client, balakrishnan2022diverse, wang2022delta, malinovsky2023federated, cho2023convergence}. These strategies are crucial as they decrease the significant variations in data quality and volume across clients~\citep{khan2021federated}. Thus, efficient client sampling is key to enhancing the performance of federated optimization.

Current sampling methodologies in FL are broadly divided into  biased~\citep{cho2020client, balakrishnan2022diverse, chen2023accelerating} and unbiased categories~\citep{el2020adaptive, wang2022delta}. Unbiased client sampling holds particular significance as it maintains the consistency of the optimization objective~\cite{wang2022delta, wang2020tackling}. 
Specifically, unlike biased sampling where client weights 
$\bs{\lambda}$  are proportional to sampling probabilities, unbiased methods separate these weights from sampling probabilities. This distinction enables unbiased sampling to be integrated effectively with strategies that address data heterogeneity~\citep{zeng2023tackling}, promote fairness~\citep{lifair, li2020tilted}, and enhance robustness~\citep{li2021ditto, li2020tilted}. Additionally, unbiased sampling aligns with secure aggregation protocols for confidentiality in FL~\citep{du2001secure, goryczka2015comprehensive, bonawitz2017practical}. Hence, unbiased client sampling techniques are indispensable for optimizing federated systems.

Therefore, a better understanding of the implications of unbiased sampling in FL could help us to design better algorithms. To this end, we summarize a general form of federated optimization algorithms with unbiased client sampling in Algorithm~\ref{alg:fl-sampler}. Despite differences in methodology, the algorithm covers unbiased sampling techniques~\citep{wang2022delta, malinovsky2023federated, cho2023convergence, salehi2017stochastic, borsos2018online, el2020adaptive, zhao2022adaptive} in the literature. In Algorithm~\ref{alg:fl-sampler}, unbiased sampling comprises three primary steps (referring to lines 3, 12, and 14). First, the \textbf{Sampling Procedure} generates a set of samples $S^t$ along with their respective probabilities. Second, the \textbf{Global Estimation} step creates global estimates for model updates, aiming to approximate the outcomes as if all participants were involved. Finally, the \textbf{Adaptive Strategy} adjusts the sampling probabilities based on the incoming information, ensuring dynamic adaptation to changing data conditions.

Typically, unbiased sampling methods in FL are founded on a \textbf{random sampling procedure}, which is then refined to improve global estimation and adaptive strategies. However, the exploration of alternative sampling procedures to enhance unbiased sampling has not been thoroughly investigated. Our research shifts focus to the \textbf{independent sampling procedure}, a less conventional approach yet viable for FL. We aim to delineate the distinctions between these methodologies as follows.
\begin{itemize}
    \item[] \textit{\textbf{Random sampling procedure (RSP)} means that the server samples clients from a black box without replacement.}
    \item[] \textit{\textbf{Independent sampling procedure (ISP)} means that the server rolls a dice for every client independently to decide whether to include the client.}
\end{itemize}

Building on the concept of arbitrary sampling~\citep{horvath2019nonconvex, chen2020optimal}, our study observes that the ISP can enhance the efficiency of estimating full participation outcomes in FL servers, as detailed in Section~\ref{sec:case_study}. However, integrating independent sampling into unbiased techniques introduces new constraints, as outlined in Remark~\ref{mk:constraints}. Addressing this innovatively in Lemma~\ref{cor:fullinformation}, our paper studies the effectiveness of general FL algorithms with adaptive unbiased client sampling, particularly emphasizing the utility and implications of the ISP from an optimization standpoint.

\begin{algorithm}[t]
\caption{FedAvg with Unbiased Client Sampler}
\label{alg:fl-sampler}
\begin{algorithmic}[1]
\REQUIRE Client set $S$, where $|S| = N$, client weights $\bs{\lambda}$, times $T$, local steps $R$
\STATE Initialize sample distribution $\bs{p}^0$ and model $\bs{x}^0$
\FOR{time $t$ in $[T]$}
    \STATE \tikzmark{l1}Server runs sampling procedure to create $S^t \sim \bs{p}^t$\tikzmark{r1}
    \STATE Server broadcasts $\bs{x}^t$ to sampled clients $i \in S^t$
    \FOR{each client $i \in S^t$ in parallel}
        \STATE $\bs{x}^{t,0}_i = \bs{x}^t$
        \FOR{local steps $r$ in $[R]$}
            \STATE $\bs{x}^{t,r}_i = \bs{x}^{t,r-1}_i - \eta_l \nabla F_i(\bs{x}^{t,r-1}_i)$
        \ENDFOR
        \STATE Client uploads local updates $\bs{g}_i^t = \bs{x}^{t,0}_i - \bs{x}^{t, R}_i$ 
    \ENDFOR
    \STATE \tikzmark{l2}Server builds estimates $\bs{d}^t = \sum_{i \in S^t} \bs{\lambda}_i \bs{g}_i^t/ \bs{p}_i^t$\tikzmark{r2}
    \STATE Server updates $\bs{x}^{t+1} = \bs{x}^t - \eta_g \bs{d}^t$
    \STATE \tikzmark{l3}Server updates $\bs{p}^{t+1}$ based on $\{\|\bs{g}_i^t\|\}_{i\in S^t}$\tikzmark{r3}
\ENDFOR
\end{algorithmic}
\drawCodeBox{fill=green}{l1}{r1}
\drawCodeBox{fill=green}{l2}{r2}
\drawCodeBox{fill=green}{l3}{r3}
\end{algorithm}

\vspace{-3.5mm}\paragraph{Contributions} 
This paper presents a comprehensive analysis of the non-convex convergence in FedAvg and its variants. We first establish a novel link between the cumulative variance of global estimates and convergence rates by separating global estimation results from heterogeneity-related factors. Thus to reduce the cumulative variance, we introduce \textsc{K-Vib}, a novel adaptive sampler incorporating the ISP. \textsc{K-Vib} notably achieves an expected regret bound of
$\tilde{\mathcal{O}}\big(N^{\frac{1}{3}}T^{\frac{2}{3}}/K^{\frac{4}{3}}\big)$, demonstrating a near-linear speed-up over existing bounds $\tilde{\mathcal{O}}\big(N^{\frac{1}{3}}T^{\frac{2}{3}}\big)$~\citep{borsos2018online} and $\mathcal{O}\big(N^{\frac{1}{3}}T^{\frac{2}{3}}\big)$~\citep{el2020adaptive}. Empirically, \textsc{K-Vib} shows accelerated convergence on standard federated tasks compared to baseline algorithms.

\section{Preliminaries}\label{sec:setup}

We first introduce previous works on batch sampling~\citep{horvath2019nonconvex} in stochastic optimization and optimal client sampling~\citep{chen2020optimal} in FL. We made a few modifications to fit our problem setup.

\begin{remark}[Constraints on sampling probability]\label{mk:constraints} \textbf{We define communication budget $K$ as the expected number of sampled clients.} And, its value range is from $1$ to $N$. To be consistent, the sampling probability $\bs{p}$ always satisfies the constraint $\bs{p}_i^t > 0, \sum_{i=1}^N \bs{p}_i^t = K, \forall t\in[T]$ in this paper.
\end{remark}

\begin{definition}[Unbiasedness of client sampling $S^t$] For communication round $t\in [T]$, the estimator $\bs{d}^t$ is related to sampling probability $\bs{p}^t$ and the sampling procedure $S^t \sim \bs{p}^t$. We define a client sampling as unbiased if the sampling $S^t$ and estimates $\bs{d}^t$ satisfy that 
$$
\mathbb{E}_{S^t}[\bs{d}^t] = \mathbb{E}\Big[\sum_{i \in S^t} \frac{\bs{\lambda}_i \bs{g}_i^t}{\bs{p}_i^t}\Big] = \sum_{i=1}^N \bs{\lambda}_i \bs{g}_i^t.$$
Besides, the variance of estimator $\bs{d}^t$ can be derived as:
\begin{equation}\label{eq:variance}
    \mathbb{V}(S^t) := \mathbb{E}_{S^t \sim \bs{p}^t}\left[\left\| \sum_{i \in S^t} \frac{\bs{\lambda}_i\bs{g}_i^t}{\bs{p}_i^t} - \sum_{i=1}^N \bs{\lambda}_i \bs{g}_i^t \right\|^2\right], 
\end{equation}
where $\mathbb{E}[|S^t|] = K$. We omit the terms $\bs{\lambda}, \bs{g}^t$ for notational brevity.
\end{definition}

\vspace{-3.5mm}\paragraph{Optimal unbiased client sampling} Optimal unbiased client sampling should achieve the lowest variance as defined in \eqref{eq:variance}. It is to estimate the global gradient of full-client participation, \ie minimize the variance of estimator $\bs{d}^t$. Given a fixed communication budget $K$, the optimum of the global estimator depends on the collaboration of sampling distribution $\bs{p}^t$ and the corresponding procedure that outputs $S^t$. 

In detail, different sampling procedures associated with the sampling distribution $\bs{p}$ build a different \textit{probability matrix} $\mathbf{P} \in \mathbb{R}^{N\times N}$, with the elements defined as $\mathbf{P}_{ij}:= \text{Prob}(\{i,j\}\subseteq S)$. Arbitrary sampling~\citep{horvath2019nonconvex} has shown the generality of denoting arbitrary sampling procedure with a probability matrix for stochastic optimization. Inspired by their findings, we focus on the optimal sampling procedure for the FL server in Lemma~\ref{lema:variance}.

\begin{lemma}[Optimal sampling procedure, \citealp{horvath2019nonconvex}]\label{lema:variance} For any communication round $t \in [T]$ in FL, random sampling yielding the $\mathbf{P}_{ij}^t = \text{Prob}(i,j\in S^t) = K(K-1)/N(N-1)$, and independent sampling yielding $\mathbf{P}_{ij}^t = \text{Prob}(i,j\in S^t) = \bs{p}_i^t \bs{p}_j^t$, they admit
\begin{equation}\label{eq:formulation}
   \mathbb{V}(S^t)  = \underbrace{\sum_{i=1}^N (1-\bs{p}_i^t) \frac{\bs{\lambda}_i^2\|\bs{g}_i^t\|^2}{\bs{p}_i^t}}_{\text{ISP}} \leq \underbrace{\frac{N-K}{N-1} \sum_{i=1}^N \frac{\bs{\lambda}_i^2\|\bs{g}_i^t\|^2}{\bs{p}_i^t}}_{\text{RSP}}.
\end{equation}
\end{lemma}
The lemma indicates that the ISP is the optimal sampling procedure that minimizes the upper bound of variance. 
Then, we have the optimal probability by solving the minimization of the upper bound in respecting probability $\bs{p}$ in Lemma~\ref{lemma:optimal}.

\begin{lemma}[Optimal sampling probability, \citealp{chen2020optimal}]\label{lemma:optimal} Generally, we can let $\bs{a}_i = \bs{\lambda}_i \|\bs{g}_i^t\|, \forall i\in[N], t\in[T]$ for simplicty of notation. Assuming $ 0 < \bs{a}_1 \leq \bs{a}_2 \leq \dots \leq \bs{a}_N$ and $0 < K \leq N$, and $l$ is the largest integer for which $0 < K+l-N \leq \frac{\sum_{i=1}^l \bs{a}_{i}}{\bs{a}_{l}}$, we have 
\begin{equation}\label{eq:optimal_p}
\begin{aligned}
\bs{p}^*_i = \begin{cases}
(K+l-N) \frac{\bs{a}_i}{\sum_{j=1}^l \bs{a}_{j}}, & \text{if}\; i \leq l, \\
1, & \text{if}\; i > l,
\end{cases}
& \quad\quad \text{(ISP)}
\end{aligned}
\end{equation}
to be a solution to the optimization problem $\min_{\bs{p}} \; \sum_{i=1}^N \frac{\bs{a}_i^2}{\bs{p}_i}$. In contrast, we provide the optimal sampling probability for the RSP
\begin{equation}
    \bs{p}^*_i = K \frac{\bs{a}_i}{\sum_{j=1}^N \bs{a}_{j}}. \quad\quad \text{(RSP)}
\end{equation}
\end{lemma}

Therefore, the optimal client sampling in FL uses ISP with probability given in \eqref{eq:optimal_p}. 

\section{Case Study on Sampling Procedure}\label{sec:case_study}

We suggest designing sampling probability for the ISP to enhance the power of unbiased client sampling in federated optimization. Beyond the tighter upper bound of variance in \eqref{eq:formulation}, three main properties can demonstrate the superiority of ISP over RSP. We illustrate the points via Example~\ref{exp:statistic} and Example~\ref{exp:illustration}.

\begin{figure}[t]
\centering
\subfigure[Sampling procedure illustration experiments\label{fig:sampling_example}]{
\begin{minipage}[t]{0.66\textwidth}
\centering
\includegraphics[width=\linewidth]{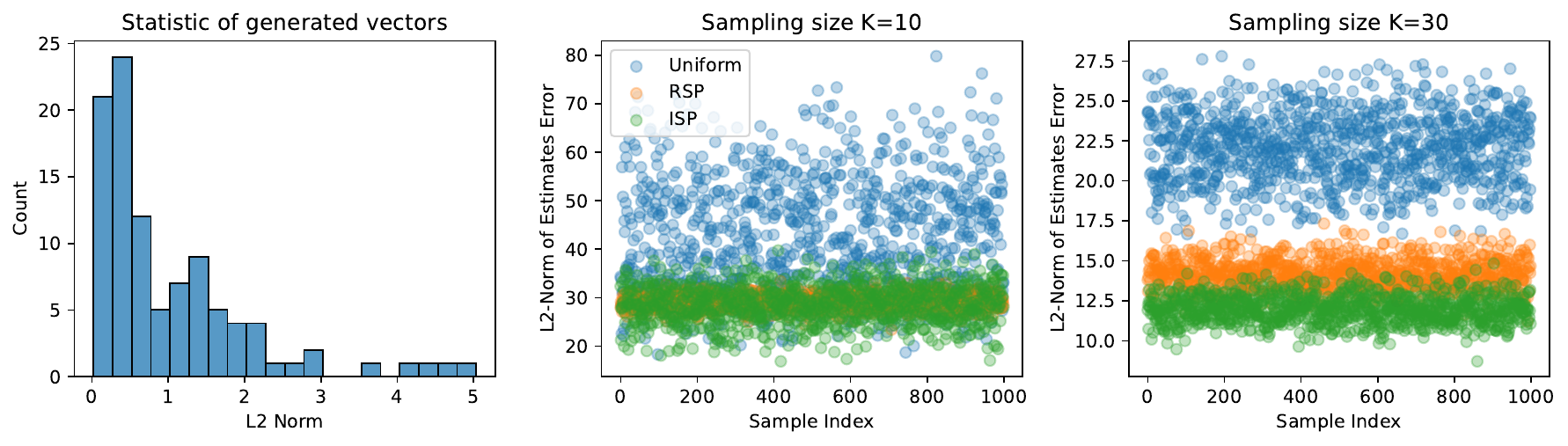}
\end{minipage}}
\hspace{0.5mm}
\subfigure[Case study\label{fig:case_study}]{
\begin{minipage}[t]{0.23\textwidth}
\includegraphics[width=\linewidth]{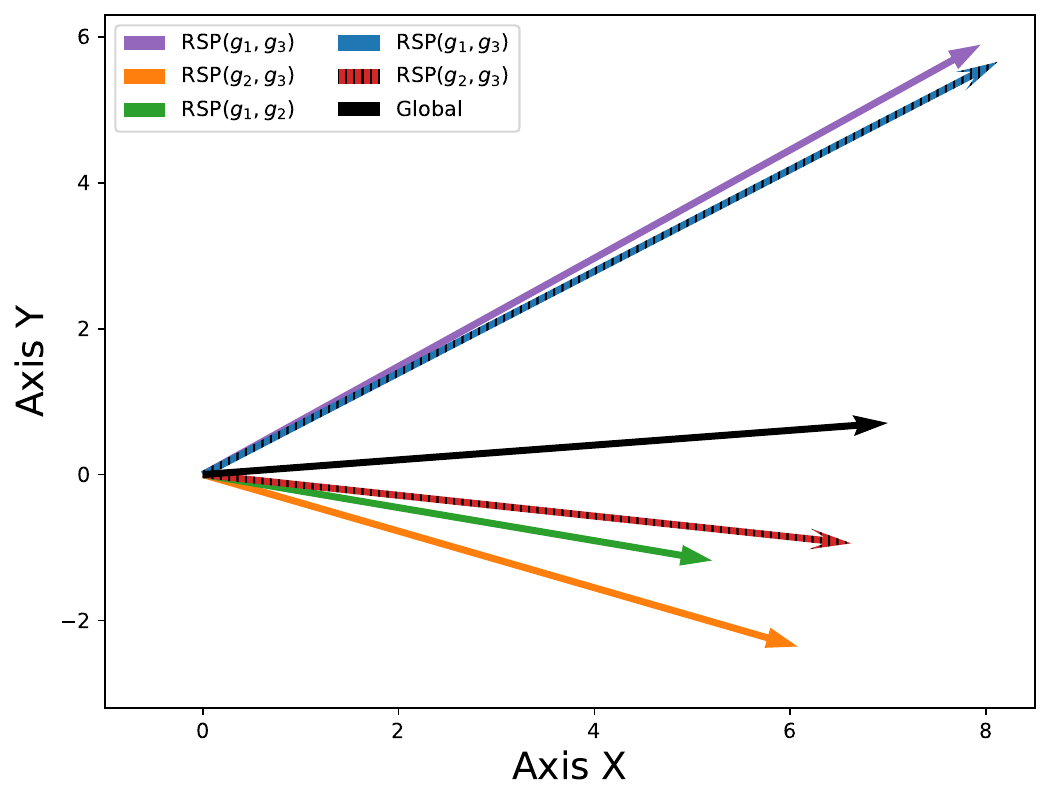}
\end{minipage}}
\hspace{0.5mm}
\caption{The variance of ISP estimates is lower than RSP. Global estimates on the X-Y plane. (a) Scatter plot of estimates errors, where ``uniform'' indicates the RSP with uniform probability. (b) The notations RSP($\bs{g}_i, \bs{g}_j$) and ISP($\bs{g}_i,\bs{g}_j$) represent the global estimates constructed through random sampling and independent sampling, respectively, using sampled vectors $\bs{g}_i$ and $\bs{g}_j$. Global indicates the full participation results. We can see ISP($\bs{g}_i,\bs{g}_j$) is closer to Global.}
\end{figure}

\begin{example}~\label{exp:statistic} We randomly generate 100 vectors with the size of 1,000 dimensions. Then, we run 1,000 times the RSP and ISP with the best probability in Lemma~\ref{lemma:optimal} to estimate full aggregation results. Then, we present the error of estimate results in Figure~\ref{fig:sampling_example}. We note the mean of these points related to \eqref{eq:variance}.
\end{example}

\begin{example}~\label{exp:illustration} Consider a case $N=3, K=2$ with $\bs{g}_1=(\frac{\sqrt{2}}{2}, \frac{\sqrt{2}}{2}), \bs{g}_2=(1, -2\sqrt{2}), g_3=(2\sqrt{7}, 2\sqrt{2})$, it induces weights vector $[\|\bs{g}_1\|^2, \|\bs{g}_2\|^2, \|\bs{g}_3\|^2] = [1,3,6]$ if omit $\bs{\lambda}$. We have optimal sampling probability $\bs{p}^* = K \cdot[0.1, 0.3, 0.6]$ for RSP and $\bs{p}^* =[0.25,0.75,1]$ for ISP with all possible estimate results in Figure~\ref{fig:case_study}.
\end{example}

\vspace{-3.5mm}\paragraph{RSP is a special case of ISP} With a minimum budget of $K=1$, the ISP does not assign any client with probability 1, it returns to the random sampling solution according to \eqref{eq:optimal_p}. If the budget $K>1$, the solution of the ISP will change, while the RSP does not. Hence, it builds better estimates than random sampling with the same sampling results as shown in the example. This is because the optimal probability of random sampling is minimizing a loose upper bound of variance \eqref{eq:formulation}. Hence, RSP tends to let each of the single estimates $\bs{a}_i/\bs{p}_i = \sum \bs{a}_i$. 

\vspace{-3.5mm}\paragraph{ISP estimates are asymptotic to full participation results} ISP builds estimates asymptotically to the full-participation results with an increasing communication budget of $K$, while random sampling does not. As shown in Figure~\ref{fig:sampling_example}, RSP and ISP achieve comparable estimates errors with lower budget $K=10$. Then, the ISP outperforms RSP with more budget $K=30$. Refer to Example~\ref{exp:illustration}, random sampling with full participation ($K=3$) builds estimates $\bs{d}^t = (6.4, 5.9)$ and hence $\|\bs{d}^t - \sum \bs{g}_i\|^2=0.6$. Analogously, full participation induces $\bs{p}^*=(1,1,1)$ for independent sampling and hence $\|\bs{d}^t - \sum \bs{g}_i\|^2=0$. 

\vspace{-3.5mm}\paragraph{ISP creates expected sampling size} The number of sampling results from independent sampling is stochastic with expectation $K$. It means that if we strictly conduct the ISP, the number of sampling results $\text{Prob}(|S^t|=K) \neq 1$, but $\mathbb{E}[|S^t|] = K$. Referring to the example, independent sampling may sample 3 clients with probability $p=0.25*0.75*1=3/16$ and sample only 1 client with $p=(1-0.25)*(1-0.75)=3/16$. Importantly, the perturbation of sampling results is acceptable due to the straggler clients~\citep{gu2021fast} in a large-scale cross-device FL system. Besides, we can easily extend our analyses to the case with straggler as discussed in Appendix~\ref{dis:fail}.

Observing the superiority of ISP, we propose to design a sampling probability and global estimates with ISP in federated learning. However, computing the optimal sampling via \eqref{eq:optimal_p} requires a norm of full gradients, which is unfeasible in practice. Therefore, FL needs a better design of its sampling probability for ISP based on limited information. In the remainder of this paper, we investigate the efficiency of ISP in federated optimization. \textbf{Unless otherwise stated, all sampling probability $\bs{p}$ and sampling procedures are related to ISP in the remainder of this paper.}

\section{General Convergence Analyses of FL with Unbiased Client Sampling}\label{sec:convergence}

In this section, we first provide a general convergence analysis of FedAvg covered by Algorithm~\ref{alg:fl-sampler}, specifically focusing on the variance of the global estimator. Our analysis aims to identify the impacts of sampling techniques on enhancing federated optimization. To this end, we define important concepts below to clarify the improvement given by an applied unbiased sampling: 

\begin{definition}[Sampling quality] Given communication budget $K$ and arbitrary unbiased client sampling probability $\bs{p}^t$, we measure the \textit{quality} (lower is better) of one sampling step $S^t\sim \bs{p}^t$ by its expectation discrepancy to the optimal sampling:
\begin{equation}\label{eq:quality}
Q(S^t) := \mathbb{E}_{S^t\sim \bs{p}^t}\left[\left\|\sum_{i\in S^t} \frac{\bs{\lambda}_i \bs{g}_i^t}{\bs{p}_i^t} - \sum_{i\in S_*^t} \frac{\bs{\lambda}_i \bs{g}_i^t}{\bs{p}_i^*}\right\|^2\right],
\end{equation}
where $S_*^t \sim \bs{p}^*$ is the ISP, $\bs{p}^*$ is obtained via \eqref{eq:optimal_p} with full $\{\|\bs{g}_i^t\|\}_{i\in[N]}$, and $\mathbb{E}[\|S^t\|] = \mathbb{E}[\|S^t_*\|] = K$. 
\end{definition}

\vspace{-3.5mm}\paragraph{Remark} Note that the second term of \eqref{eq:quality} denotes the best results that can be possibly obtained subjected to communication budget. It still preserves estimate errors to full results. Therefore, we define the sampling quality of one sampling by its gap to the optimal estimate results for practical concern.



In practical settings, federated learning typically trains modern neural networks, which are non-convex problems in optimization. Therefore, our convergence analyses rely on standard assumptions on the local empirical function $f_i, i\in[N]$ in non-convex federated optimization~\citep{chen2020optimal, jhunjhunwala2022fedvarp, chen2023accelerating}.

\begin{assumption}[Smoothness]\label{asp:smoothness}
    Each objective $f_i(\bs{x})$ for all $i\in[N]$ is $L$-smooth, inducing that for all $\forall \bs{x}, \bs{y}\in\mathbb{R}^d$, it holds $\|\nabla f_i(\bs{x}) - \nabla f_i(\bs{y})\| \leq L \|\bs{x} - \bs{y}\|$.
\end{assumption}
\begin{assumption}[Unbiasedness and bounded local variance]\label{asp:unbiasedness}
    For each $i\in[N]$ and $\bs{x} \in \mathbb{R}^d$, we assume the access to an unbiased stochastic gradient $\nabla F_i(\bs{x}, \xi_i)$ of client's true gradient $\nabla f_i(\bs{x})$, i.e.,$\mathbb{E}_{\xi_i\sim\mathcal{D}_i}\left[\nabla F_i(\bs{x}, \xi_i)\right] =\nabla f_i(\bs{x})$.  The function $f_i$ have $\sigma_l$-bounded (local) variance i.e.,$\mathbb{E}_{\xi_i\sim\mathcal{D}_i}\left[\left\|\nabla F_i(\bs{x}, \xi_i)-\nabla f_i(\bs{x})\right\|^2\right] \leq \sigma_l^2$.
\end{assumption}

\begin{assumption}[Bounded global variance]\label{asp:bounded-variance} 
We assume the weight-averaged global variance is bounded, i.e., $\sum_{i=1}^N\bs{\lambda}_i\left\|\nabla f_i(\bs{x})-\nabla f(\bs{x})\right\|^2 \leq \sigma_{g}^2$ for all $\bs{x} \in \mathbb{R}^d$.
\end{assumption}

Assumptions~\ref {asp:smoothness} and~\ref{asp:unbiasedness} are standard assumptions in stochastic optimization analyses. Assumption~\ref{asp:bounded-variance} measures the impacts of data heterogeneity in federated optimization. A larger upper bound of $\sigma_g^2$ denotes stronger heterogeneity across clients. Now, we provide the non-convex convergence of Algorithm~\ref{alg:fl-sampler}.

\begin{theorem}[FedAvg with arbitrary unbiased client sampling]\label{theorem:convergence} Under Assumptions~\ref{asp:smoothness},~\ref{asp:unbiasedness},~\ref{asp:bounded-variance}, taking upper bound $\mathbb{E}\left[f(\bs{x}^0)-f(\bs{x}^{+\infty})\right] \leq M$ and $W = \max \{\bs{\lambda}_i\}_{i\in[N]}$, given communication budget $K$, there always exists learning rates $\eta_l \eta_g \leq \frac{1}{8RL}$ allow Algorithm~\ref{alg:fl-sampler} to generate an iteration sequence $\{\bs{x}^1, \dots, \bs{x}^t\}$ such that
\begin{equation}\label{eq:conv_speed}
\begin{aligned}
\min_{t\in[T]} \mathbb{E} \|\nabla f(\bs{x}^t)\|^2 \leq \mathcal{O}\left(\frac{ML\sigma^2}{T}\right)^{\frac{1}{2}} + \underbrace{\mathcal{O}\left(\frac{M\beta_1}{T}\right)^{\frac{1}{2}} + \mathcal{O}\left(\frac{M(\sigma^2 \beta_2)^2}{T}\right)^{\frac{2}{3}}}_{\text{Sampling Utility}} + \mathcal{O}\left(\frac{ML}{T}\right).
\end{aligned}
\end{equation}
where
$$
\begin{aligned}
\sigma^2 = \sigma_l^2/R + \sigma_g^2, \quad \beta_1 = \frac{1}{T}\sum_{t=0}^{T-1} Q(S^t), \quad \beta_2 = \frac{1}{T}\sum_{t=0}^{T-1} \gamma^t, \quad \gamma^t \in [NW, NW + \frac{N-K}{K}W].
\end{aligned}
$$
Notably, $\sigma^2$ denotes the heterogeneity impacts on stochastic gradients, $\beta_2$ denotes the benefits of using optimal sampling, and $\beta_1$ denotes the benefits of using sub-optimal sampling.
\end{theorem}

\vspace{-3.5mm}\paragraph{Interpretation of Theorem~\ref{theorem:convergence}} 
The convergence guarantees quantify the impacts of client sampling quality on the convergence performance of FedAvg. If we always use optimal client sampling~\citep{chen2020optimal}, the cumulative sampling quality $\beta_1 = 0$. Then, the convergence rate returns to Theorem 18~\citep{chen2020optimal} and covers the state-of-art complexity guarantees in \cite{karimireddy2020scaffold}. As discussed previously, acquiring optimal client sampling is typically unfeasible in FL practice. Therefore, we may use sub-optimal client sampling techniques, which induce $\beta_1$ as a non-zero value.

\vspace{-3.5mm}\paragraph{Sampling utility} 
Given communication budget $K$, the utility of optimal client sampling is linked to $\gamma^t \in [NW, NW + \frac{N-K}{K}W]$, where $W = \max \{\bs{\lambda}_i\}_{i\in[N]}$. For example, $\gamma^t = NW$ indicates the best case that optimal client sampling can accurately approximate full results. Otherwise, the optimal client sampling implements sub-optimal approximation. Moreover, given an arbitrary sub-optimal client sampling strategy, we use $Q(S^t)$ to measure the discrepancy between the applied sampling and optimal sampling. Using limited information when optimal client sampling is inapplicable, we expect to minimize the cumulative discrepancies $\beta_1$ to obtain optimization improvement.
In all, the bound of the second term in \eqref{eq:conv_speed} is related to the performance of the applied client sampler in FL. Therefore, minimizing the cumulative sampling quality $\beta_1$ over federated optimization iteration directly accelerates the FL convergence.

\section{Theories of the K-Vib Sampler}\label{sec:kvib_sampler}

In this section, we introduce the theoretical design of the K-Vib sampler for federated client sampling. The adaptive sampling objective aligns with the online variance reduction~\citep{salehi2017stochastic, borsos2018online, el2020adaptive} tasks in stochastic optimization. The difference is that we solve the problem in the scenario of FL using ISP, which induces the constraints on sampling probability given in Remark~\ref{mk:constraints}.

\subsection{Adaptive Client Sampling as Online Optimization}

To enhance federated optimization, we aim to minimize the sampling quality $Q(S^t)$ to achieve tighter convergence bound \eqref{eq:conv_speed}. And, using ISP variance, the upper bound of \eqref{eq:quality} can be known as:
$$
\begin{aligned}
Q(S^t) & \leq \sum_{i=1}^N \frac{\bs{\lambda}_i^2\|\bs{g}_i^t\|^2}{\bs{p}_i^t} - \sum_{i=1}^N \frac{\bs{\lambda}_i^2\|\bs{g}_i^t\|^2}{\bs{p}_i^*}.
\end{aligned}
$$
Then, we model the client sampling objective as an online convex optimization problem~\citep{salehi2017stochastic, borsos2018online, el2020adaptive}. Concretely, we define 
$$
\text{local feedback function}\; \pi_t(i):=\bs{\lambda}_i \|\bs{g}_i^t\|, \;\text{and cost function}\; \ell_t(\bs{p}):= \sum_{i=1}^N \frac{\pi_t(i)^2}{\bs{p}_i}
$$ 
for a online convex optimization task\footnote{Please distinguish the online cost function $\ell_t(\cdot)$ from local empirical loss of client $f_i(\cdot)$ and global loss function $f(\cdot)$. While $\ell_t(\cdot)$ is always convex, $f(\cdot)$ and $f_i(\cdot)$ can be non-convex.} respecting sampling probability $\bs{p}$. Therefore, online convex optimization is to minimize a \textit{dynamic} regret defined as:
\begin{equation}\label{eq:regret}
\begin{aligned}
    \frac{1}{T}\sum_{t=1}^T Q(S^t) & \leq \frac{1}{T}\text{Regret}_D(T) := \frac{1}{T}\left(\sum_{t=1}^T \ell_t(\bs{p}^t) - \sum_{t=1}^T \min_{\bs{p}} \ell_t(\bs{p})\right).
\end{aligned}
\end{equation} 

\vspace{-3.5mm}\paragraph{What does \textit{regret} measure?} 
\textit{Regret} measures the cumulative discrepancy of applied sampling probability and the \textit{dynamic} optimal Oracle. In Theorem~\ref{theorem:convergence}, we decomposed the cumulative sampling quality as an error term. And, the upper bound of cumulative sampling quality is given by the \textit{regret}. According to \eqref{eq:formulation}, the ISP induces a tighter regret. Minimizing the regret \eqref{eq:regret} can devise sampling probability for ISP to create a tighter convergence rate for applied FL.
In this paper, we are to build an efficient sampler that outputs an exemplary sequence of independent sampling distributions $\{\bs{p}^t\}_{t=1}^T$ such that $\lim_{T\rightarrow \infty} \text{Regret}_D(T)/T = 0$. 

\subsection{Analyzing the Best Fixed Probability}

In the federated optimization process, the local updates $\bs{g}^t$ change, making it challenging to directly bound the cumulative discrepancy between the sampling probability and the dynamic optimal probability. Consequently, we explore the advantages of employing the best-fixed probability instead.
We decompose the \eqref{eq:regret} into:
\begin{equation}\label{eq:regret-dec}
\begin{aligned}
     \text{Regret}_D(T) & = \underbrace{\sum_{t=1}^T \ell_t(\bs{p}^t) - \min_{\bs{p}}\sum_{t=1}^T \ell_t(\bs{p})}_{\text{Regret}_S(T)} + \underbrace{\min_{\bs{p}}\sum_{t=1}^T \ell_t(\bs{p}) -  \sum_{t=1}^T \min_{\bs{p}} \ell_t(\bs{p})}_{T_\text{BFP}}.
\end{aligned}
\end{equation}
The static regret $\text{Regret}_S(T)$ denotes the cumulative online loss gap between an applied sequence of probabilities and the \textit{best-fixed} probability in hindsight. The second term indicates the cumulative loss gap between the best-fixed probability in hindsight and the optimal probabilities. We are to bound the terms respectively.

Our analyses rely on a mild assumption of the convergence status of the federated optimization that sampling methods are applied~\citep{wang2021field}. 
Notably, stochastic optimization~\citep{salehi2017stochastic, duchi2011adaptive, boyd2004convex} and federated optimization algorithms~\citep{reddi2020adaptive, wang2020tackling, li2019convergence} typically achieve a sub-linear convergence speed $\mathcal{O}(1/\sqrt{T})$ at least. Therefore, we assume feedback function related to the convergence behaviors of local objectives $f_i(\cdot), i\in[N]$ using the following notions:
\vspace{-3.5mm}\paragraph{Notions} We denote the overall feedback $\Pi_t := \sum_{i=1}^N \pi_t(i)$ at $t$-th round. Then, we denote the local convergence results $\pi_*(i) := \lim_{t\rightarrow \infty} \pi_t(i)$, and the overall convergence results $\Pi_* := \sum_{i=1}^N \pi_*(i), \; \forall i \in [N]$. Notably, we know $\sum_{t=1}^T \Pi_t \geq \Pi_*$ as FL converges and we denote $V_T(i) = \sum_{t=1}^T \big(\pi_t(i) - \pi_*(i)\big)^2, \forall T \geq 1,$. Besides, we denote the largest feedback with $G$, i.e., $\pi_t(i) \leq G, \forall t\in[T], i \in[N]$. Importantly, the $G$ denotes the largest feedback during the applied optimization process, instead of assuming bounded gradients. 

Then, we assume the convergence of federated optimization will induce a decaying speed of feedback function (local update norms $\|\boldsymbol{g}\|$ in this paper):
\begin{assumption}[Convergence of applied federated optimization]\label{asp:local_convergence}

As we discussed above the sub-linear convergence speed $\mathcal{O}(1/\sqrt{T})$ can be obtained by general nonconvex federated learning algorithms. \textit{We assume that $|\pi_t(i) - \pi_*(i)| \leq \mathcal{O}(1/\sqrt{t})$, and hence implies $V_T(i) \leq \mathcal{O}(\log(T))$}. 
The above assumptions guarantee the regret concerning a basic convergence speed of applied FL algorithms, with an additional cost of $\tilde{\mathcal{O}}(\sqrt{T})$. 
\end{assumption}

Now, we bound the second term of \eqref{eq:regret-dec} below:
\begin{theorem}[Bound of best fixed probability]\label{bound:hindsight} Under Assumptions~\ref{asp:local_convergence}, sampling a set of clients with an expected size of $K$, and for any $i \in [N]$ denote $V_T(i) = \sum_{t=1}^T \big(\pi_t(i) - \pi_*(i)\big)^2 \leq \mathcal{O}(\log(T))$. For any $T \geq 1$, the averaged hindsight gap admits,
\begin{equation}\nonumber
T_\text{BFP} \leq \frac{T}{K} \left(\sum_{i=1}^N \sqrt{\frac{V_T(i)}{T}} \right)\left(2 \Pi_* + \sum_{i=1}^N \sqrt{\frac{V_T(i)}{T}}\right).
\end{equation}
\begin{proofsketch}
This bound can be directly proved to solve the convex optimization problem respectively. Please see Appendix~\ref{proof:bound:hindsight} for details.
\end{proofsketch}
\end{theorem}

Theorem~\ref{bound:hindsight} indicates a fast convergence of federated optimization induces a better bound of $V_T(i)$, yielding a tighter regret. Therefore, it also covers better optimization problems implying a tighter upper bound assumption for $|\pi_t(i) - \pi_*(i)|$ (e.g., strongly convex and convex federated learning problems). 
As the hindsight bound vanishes with an appropriate FL solver, our objective turns to devise a $\{\bs{p}_1, \dots, \bs{p}_T\}$ that bounds the static regret $\text{Regret}_S(T)$ in \eqref{eq:regret-dec}.

\subsection{Upper Bound of Static Regret}

We utilize the classic follow-the-regularized-leader (FTRL)~\citep{shalev2012online, kalai2005efficient, hazan201210} framework to design a stable sampling probability sequence, which is formed at time $t$:
\begin{equation}\label{obj:ol_ftrl}
    \bs{p}^t := \arg \min_{\bs{p}} \left\{\sum_{i=1}^N \frac{ \pi_{1:t-1}^2(i) + \gamma}{\bs{p}_i}\right\},
\end{equation}
where the regularizer $\gamma$ ensures that the probability does not change too much and prevents assigning a vanishing probability to clients. It also ensures a minimum sampling probability $p_{\text{min}}$ for some clients. Therefore, we have the closed-form solution as shown below:
\begin{lemma}[Solution to \eqref{obj:ol_ftrl}]\label{cor:fullinformation} Letting $\bs{a}_i^t = \pi_{1:t-1}^2(i) + \gamma$ and $ 0 < \bs{a}_1^t \leq \bs{a}_2^t \leq \dots \leq \bs{a}_N^t$ and $0 < K \leq N$, we have 
\begin{equation}\label{eq:time_results_p}
    \bs{p}_i^t = \begin{cases}
        1, & \text{if}\; i\geq l_2 ,\\
        z_t \frac{\sqrt{\bs{a}_i^t}}{c_t}, & \text{if}\; i \in (l_1, l_2), \\
        p_{\text{min}}, & \text{if}\; i\leq l_1, \\
    \end{cases}
\end{equation}
where $c_t=\sum_{i\in(l_1,l_2)} \sqrt{\bs{a}_i^t}$, $z_t = K - (N-l_2)+l_1\cdot p_{\text{min}}$ and the $1 \leq l_1 \leq l_2 \leq N$, which satisfies that $\forall i \in (l_1, l_2)$,
$$
\frac{p_{\text{min}} \cdot \sum_{l_1 < i < l_2} \bs{a}_i^t}{z_t} < \bs{a}_i^t < \frac{\sum_{l_1 < i < l_2} \bs{a}_i^t}{z_t}.
$$
\end{lemma}

\textbf{Remark.} Compared with vanilla optimal sampling probability in \eqref{eq:optimal_p}, our sampling probability especially guarantees a minimum sampling probability $p_{\text{min}}$ on the clients with lower feedback. This probability encourages the exploration of the FL system and prevents the case that some clients are never sampled. Besides, the minimum sampling probability $p_{\text{min}}$ is determined by the $\gamma$ and the cumulative feedback from clients during training. For $t = 1,\dots,T$, if applied sampling probability follows Lemma~\ref{cor:fullinformation} with a proper $\gamma$, we guarantee that $\text{Regret}_S(T)/T \leq \mathcal{O}(1/\sqrt{T})$, as proved in Appendix~\ref{apd:full_info}. 

However, under practical constraints, the server only has access to the feedback information from the past sampled clients. Hence, \eqref{eq:time_results_p} can not be computed accurately. Inspired by \citealp{borsos2018online}, we construct an additional estimate of the true feedback for all clients denoted by $\tilde{\bs{p}}$ and let $S^t \sim \tilde{\bs{p}}^t$. Concretely, $\tilde{\bs{p}}$ is mixed by the original estimator $\bs{p}^t$ with a static distribution. Let $\theta \in [0,1]$, we have
\begin{equation}\label{eq:utilize}
\text{Mixing strategy:}\quad\quad \tilde{\bs{p}}^t = (1-\theta)\bs{p}^t + \theta \frac{K}{N},
\end{equation}
where $\tilde{\bs{p}}^t \geq \theta \frac{K}{N}$, and hence $\tilde{\pi}^2_t(i) \leq \pi^2_t(i)\cdot\frac{N}{\theta K} \leq G^2\cdot\frac{N}{\theta K}$. 

Analogous to regularizer $\gamma$, the mixing strategy guarantees the least probability that any clients be sampled, thereby encouraging exploration. 
We present the expected regret bound of the sampling with mixed probability and the K-Vib sampler outlined in Algorithm~\ref{alg:sampler} with theoretical guarantee in Theorem~\ref{bound:soft-regret}.
\begin{theorem}[Static expected regret with partial feedback]\label{bound:soft-regret} Under Assumptions~\ref{asp:local_convergence}, sampling $S^t \sim \tilde{\bs{p}}^t$ with $\mathbb{E}[|S^t|]=K$ for all $t=1,\dots,T$, and letting $\gamma=G^2 \frac{N}{K\theta}, \theta = (\frac{N}{TK})^{1/3}$ with $T\cdot K \geq N$, we obtain the expected regret
\begin{equation}\label{eq:final_regret}
    \mathbb{E}\left[\text{Regret}_S(T)\right] \leq \tilde{\mathcal{O}}\big(N^{\frac{1}{3}}T^{\frac{2}{3}}/K^{\frac{4}{3}}\big),
\end{equation}
where $\tilde{\mathcal{O}}$ hides the logarithmic factors. 
\begin{proofsketch}
Denoting $\{\pi_t(i)\}_{i \in S^t}$ as partial feedback from sampled points, it incurs
\begin{equation}\nonumber
\tilde{\pi}^2_t(i) := \frac{\pi^2_t(i)}{\tilde{\bs{p}}^t_i}\cdot \mathbb{I}_{i\in{S^t}}, \text{and} \; \mathbb{E}[\tilde{\pi}^2_t(i) | \tilde{\bs{p}}^t_i] = \pi^2_t(i), \forall i \in [N].
\end{equation}
Analogous to \eqref{eq:regret}, we define modified cost functions and their unbiased estimates:
$$
\tilde{\ell}_t(\bs{p}):= \sum_{i=1}^N \frac{\tilde{\pi}^2_t(i)}{\bs{p}_i}, \text{and} \; \mathbb{E}[\tilde{\ell}_t(\bs{p}) | \tilde{\bs{p}}^t, \ell_t ] = \ell_t(\bs{p}).
$$ 

Our hyperparameters $\gamma, \theta$ are independent. The $\gamma$ is set to guarantee the stability of probability sequence in \eqref{bound:ftrl-al}. The $\theta$ is set to optimize the final upper bound. Relying on the additional estimates, we have the full cumulative feedback in expectation. In detail, we provide regret bound $\text{Regret}_S(T)$ by directly using Lemma~\ref{cor:fullinformation} in Appendix~\ref{apd:full_info}. Analogously, we can extend the mixed sampling probability $\tilde{\bs{p}}^t$ to derive the expected regret bound $\mathbb{E}[\text{Regret}_S(T)]$ given in Appendix~\ref{proof:bandit}.
\end{proofsketch}
\end{theorem}

\begin{algorithm}[t]
\caption{K-Vib Sampler}
\label{alg:sampler}
\begin{algorithmic}
\renewcommand{\COMMENT}[2][.6\linewidth]{%
  \leavevmode\hfill\makebox[#1][l]{$\triangleright$~#2}}
\REQUIRE $N$, $K$, $T$, $\gamma$, and $\theta$.
\STATE Initialize client feedback storage $\omega(i)=0$ for all $i\in[N]$.
\FOR{time $t$ in $[T]$}
    \STATE $\bs{p}_i^t \propto \sqrt{\omega(i) + \gamma}$ \COMMENT{by Lemma~\ref{cor:fullinformation}}
    \STATE $\tilde{\bs{p}}^t_i \leftarrow (1-\theta)\cdot \bs{p}_i^t + \theta \frac{K}{N}$, for all $i \in [N]$
    \STATE Draw $S^t \sim \tilde{\bs{p}}^t$ \COMMENT{ISP}
    \STATE Receive feedbacks $\pi_t(i)$, and update $\omega(i) \leftarrow \omega(i) + \pi_t^2(i)/\tilde{\bs{p}}^t_i$ for $i \in S^t$
\ENDFOR
\end{algorithmic}
\end{algorithm}

\vspace{-3.5mm}\paragraph{Enhanced convergence rate of FedAvg with K-Vib sampler}
The K-Vib sampler can work with a federated optimization process providing unbiased full result estimates. Comparing with previous regret bound $\tilde{\mathcal{O}}\big(N^{\frac{1}{3}}T^{\frac{2}{3}}\big)$~\citep{borsos2018online} and $\mathcal{O}\big(N^{\frac{1}{3}}T^{\frac{2}{3}}\big)$~\citep{el2020adaptive}, it implements a linear speed up with communication budget $K$. This advantage relies on a tighter formulation of variance obtained via the ISP.
Furthermore, the Assumption~\ref{asp:local_convergence} holds in FedAvg. Meanwhile, the regret bound of the K-Vib sampler is independent of the convergence of Algorithm~\ref{alg:fl-sampler}. Therefore, the K-Vib sampler can accelerate FedAvg by minimizing the second term in Theorem~\ref{theorem:convergence}. Concretely, the upper bound of dynamic regret in~\eqref{eq:regret-dec} is dominated by static regret from Theorem~\ref{bound:soft-regret}. And then, we combine \eqref{eq:regret} and substitute $\beta_1$ in \eqref{eq:conv_speed}, which derives that \textbf{K-Vib sampler improves the second term $\mathcal{O}(\sqrt{M\beta_1/T})$ in Theorem~\ref{theorem:convergence} to $\mathcal{O}\left(\frac{M^{3/4} N^{1/4} }{TK}\right)^{\frac{2}{3}}$.} For computational complexity, the primary cost involves sorting the cumulative feedback sequence $\{\omega(i)\}_{i=1}^N$ in Algorithm~\ref{alg:sampler}. This sorting operation can be performed efficiently with an adaptive sorting algorithm~\citep{estivill1992survey}, resulting in a time complexity of at most $\mathcal{O}(N \log N)$. And, we provide a sketch of efficient implementation in Appendix~\ref{sec:efficient}.

\section{Experiments}\label{sec:exp}

This section evaluates the convergence benefits of utilizing FL client samplers. Our experiment evaluation aligns with previous works~\citep{li2020federated, chen2020optimal} on synthetic and Federated EMNIST datasets. And, we additionally evaluate our method on language model and text datasets.

\vspace{-3.5mm}\paragraph{Baselines} 
We compare the K-Vib sampler with the uniform sampling and other adaptive unbiased samplers including Multi-armed Bandit Sampler (Mabs)~\citep{salehi2017stochastic}, Variance Reducer Bandit (Vrb)~\citep{borsos2018online} and Avare~\citep{el2020adaptive}. We run experiments with the same random seed and vary the seeds across five independent runs. We present the mean performance with the standard deviation (error bars). To ensure a fair comparison, we set the hyperparameters of all samplers to the optimal values prescribed in the original papers, and concrete hyperparameters are detailed in Appendix~\ref{app:experiments}. 

\vspace{-3.5mm}\paragraph{FL and sampler hyperparameters} 
We run $T=500$ round for all tasks and use vanilla SGD optimizers with constant step size for both clients and the server. We fix $\eta_g = 1$ and tune $\eta_l$ for different tasks. 
For K-Vib sampler, we set $\theta = (\frac{N}{TK})^{\frac{1}{3}}$, which aligns with Theorem~\ref{bound:soft-regret}. Then, we guarantee the stability of designed probability via setting $\gamma \approx G^2\frac{N}{\theta K}$. In practice, we suggest using the mean value of first-round client feedback as a naive estimate of $G$. The first-round feedback is typically the largest during FL training based on Assumption~\ref{asp:local_convergence}. This hyperparameter tuning experience can be applied in future applications.

\begin{figure*}
\centering
\includegraphics[width=0.85\linewidth]{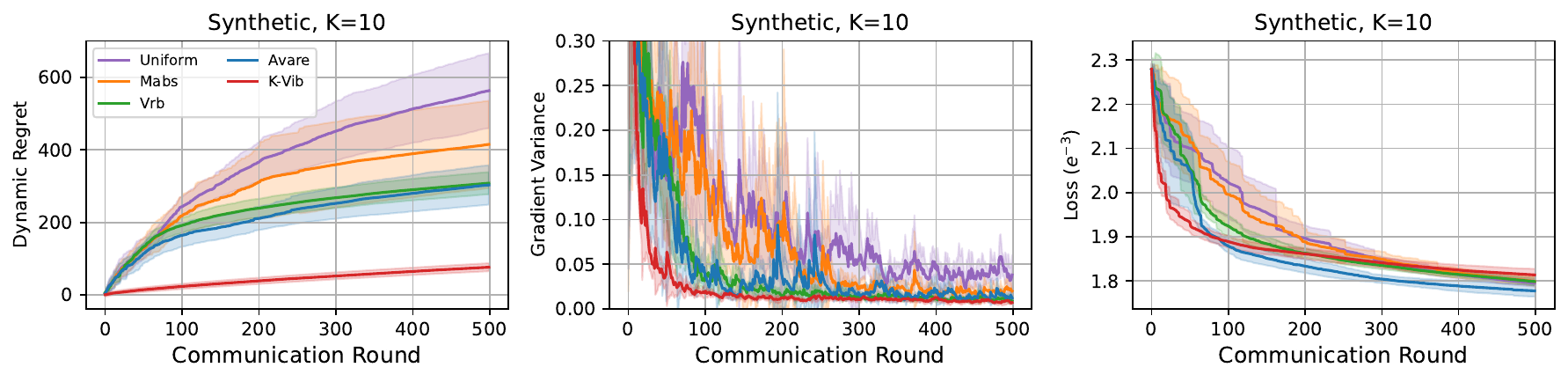}
\vspace{-4mm}
\caption{Evaluation on dynamic regret~\eqref{eq:regret}, gradient variance~\eqref{eq:variance}, and test loss.\label{fig:online_metric}}
\vspace{-5mm}
\end{figure*}

\begin{figure*}[t]
\centering
\subfigure[Synthetic dataset\label{fig:synthetic_dist}]{
\begin{minipage}[t]{0.22\textwidth}
\centering
\includegraphics[width=\linewidth]{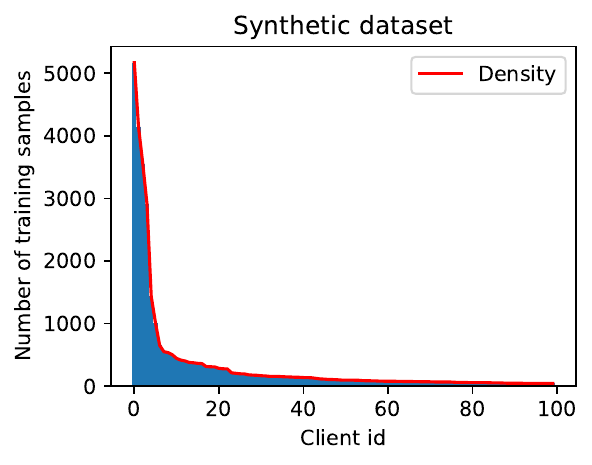}
\end{minipage}}
\subfigure[Regret improvement.\label{fig:effct_k}]{
\begin{minipage}[t]{0.23\textwidth}
\centering
\includegraphics[width=\linewidth]{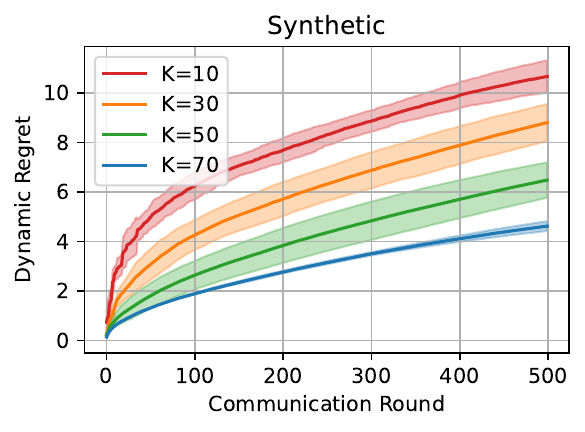}
\end{minipage}}
\subfigure[Sensitivity to regularization $\gamma$.\label{fig:sensitivity}]{
\begin{minipage}[t]{0.45\textwidth}
\centering
\includegraphics[width=\linewidth]{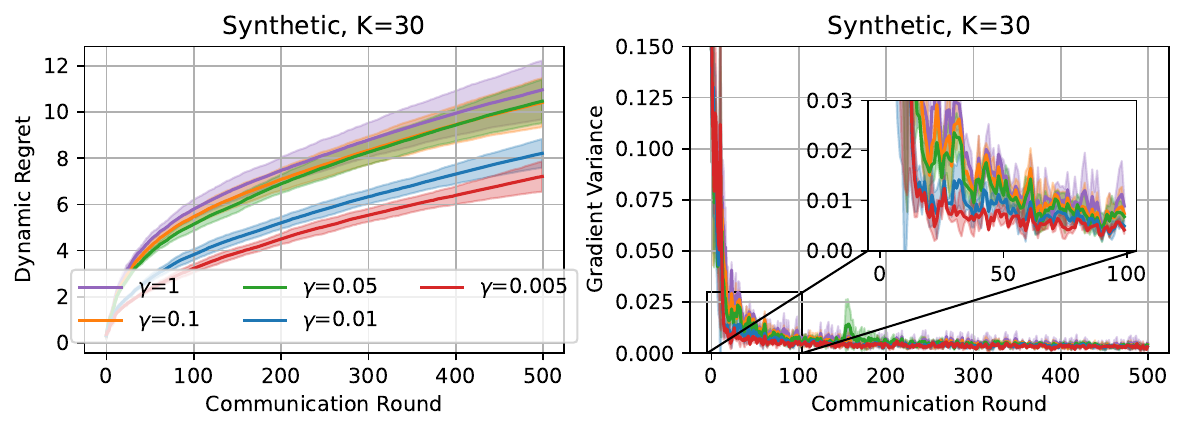}
\end{minipage}}
\caption{Data distribution of synthetic dataset and sensitivity study on $\gamma$.}
\end{figure*}

\subsection{Synthetic Dataset}

We evaluate the theoretical results via experiments on Synthetic datasets, where the data are generated from Gaussian distributions~\citep{li2020federated} and the model is logistic regression $f(\bs{x})=\arg\max(W^T \bs{x} + \bs{b})$. We generate $N=100$ clients of each has a synthetic dataset, where the size of each dataset follows the power law as shown in Figure~\ref{fig:synthetic_dist}. We set local learning rate $\eta_l = 0.02$, local epoch 1, and batch size 64.

In Figure~\ref{fig:online_metric}, we show the action of all samplers on three metrics. Concretely, the K-Vib implements a lower curve of regret in comparison with baselines. Hence, it creates a better estimate with lower variance for global model updating. Connecting with Theorem~\ref{theorem:convergence}, FedAvg with K-Vib achieves a faster convergence. 

We present Figure~\ref{fig:effct_k} to prove the linear speed up about communication budget $K$ in Theorem~\ref{bound:soft-regret}. In detail, with the increase of budget $K$, the performance of the K-Vib sampler with regret metric is reduced significantly. Due to page limitation, we provide further illustration examples of other baselines in the same metric in Appendix Figure~\ref{fig:baseline_effect_k}, where we show that the regret bound of baselines methods are not reduced with increasing communication budget $K$. The results demonstrate our unique improvements in theories.

Figure~\ref{fig:sensitivity} reveals the effects of regularization $\gamma$ in Algorithm~\ref{alg:sampler}. The regret slightly changes with different $\gamma$. The variance reduction curves remain stable, indicating the K-Vib sampler is not sensitive to $\gamma$. This is because the regularizer $\gamma$ only decides the minimum probability in solution~\eqref{eq:time_results_p}. 

\subsection{Federated EMNIST Dataset}
\begin{figure*}[t]
\centering
\subfigure[Distribution of Federated EMNIST dataset\label{fig:femnist_dist}]{
\begin{minipage}[t]{0.8\textwidth}
\centering
\includegraphics[width=\linewidth]{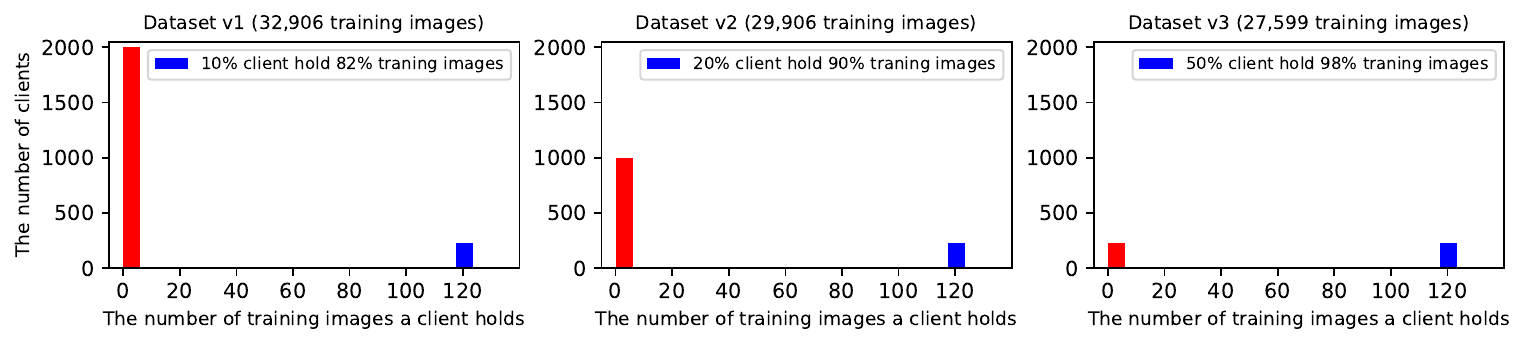}
\end{minipage}}
\hspace{0.5mm}
\subfigure[Training loss and test accuracy of FedAvg with different unbiased samplers.\label{fig:femnist_optimization}]{
\begin{minipage}[t]{0.8\textwidth}
\includegraphics[width=\linewidth]{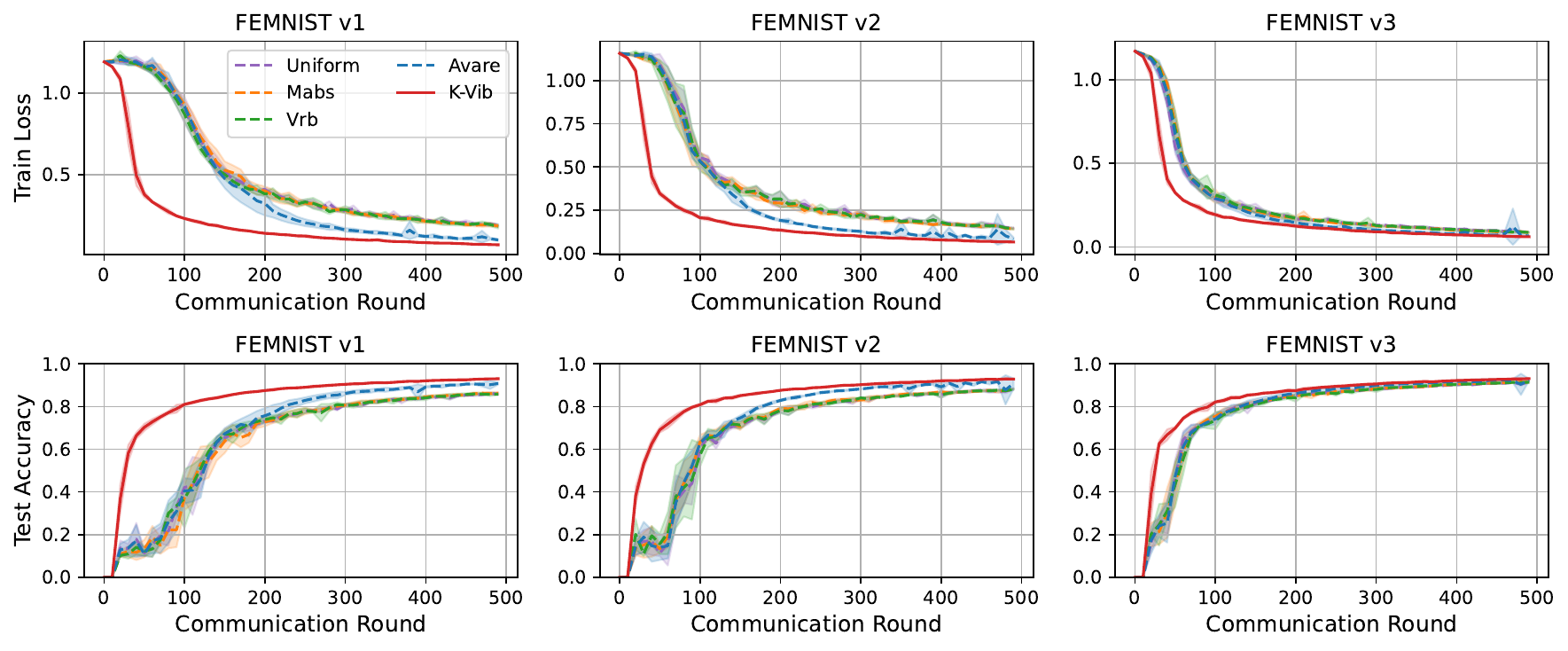}
\end{minipage}}
\hspace{0.5mm}
\caption{Federated EMNIST dataset experiments.}\label{fig:dataset_dist}
\end{figure*}

We evaluate the proposed sampler on the Federated EMNIST (FEMNIST) following~\citealp{chen2020optimal} for image classification. The FEMNIST tasks involve three degrees of unbalanced level~\citep{chen2020optimal} as shown in Figure~\ref{fig:femnist_dist}, including FEMNIST v1 (2,231 clients in total, 10\% clients hold 82\% training images), FEMNIST v2 (1,231 clients in total, 20\% client hold 90\% training images) and FEMNIST v3 (462 clients in total, 50\% client hold 98\% training images). We use the same CNN model in ~\citep{mcmahan2017communication}. We set batch size 20, local epochs 3, $\eta_l = 0.01$, and $K=111, 62, 23$ as 5\% of total clients. 

In Figure~\ref{fig:femnist_optimization}, the variance of data quantity decreased from FEMNIST v1 to FEMNIST v3. We observe that the FedAvg with the K-Vib sampler converges about 3$\times$ faster than baseline when achieving 75\% accuracy in FEMINIST v1 and 2$\times$ faster in FEMINIST v2. At early rounds, the global estimates provided by naive independent sampling are better as demonstrated in Lemma~\ref{lema:variance}, it induces faster convergence by Theorem~\ref{theorem:convergence}. Meanwhile, the K-Vib sampler further enlarges the convergence benefits by solving an online variance reduction task. Hence, it maintains a fast convergence speed. For baseline methods, we observe that the Vrb and Mabs do not outperform the uniform sampling in the FEMNIST task due to the large number of clients and large data quantity variance. In contrast, the Avare sampler fastens the convergence curve after about 150 rounds of exploration in the FEMNIST v1 and v2 tasks. On the FEMNIST v3 task, the Avare sampler shows no clear improvement in the convergence curve, while the K-Vib sampler still implements marginal improvements. Horizontally comparing the results, we observe that the curve discrepancy between K-Vib and baselines is the largest in FEMNIST v1. And, the discrepancy narrows with the decrease of data variance across clients.  It indicates that the K-Vib sampler works better in the cross-device FL system with a large number of clients and data variance.

\begin{figure*}[t]
\centering
\subfigure[Experiments on AGNews dataset with DistillBert model.\label{fig:agnews}]{
\begin{minipage}[t]{0.8\textwidth}
\centering
\includegraphics[width=\linewidth]{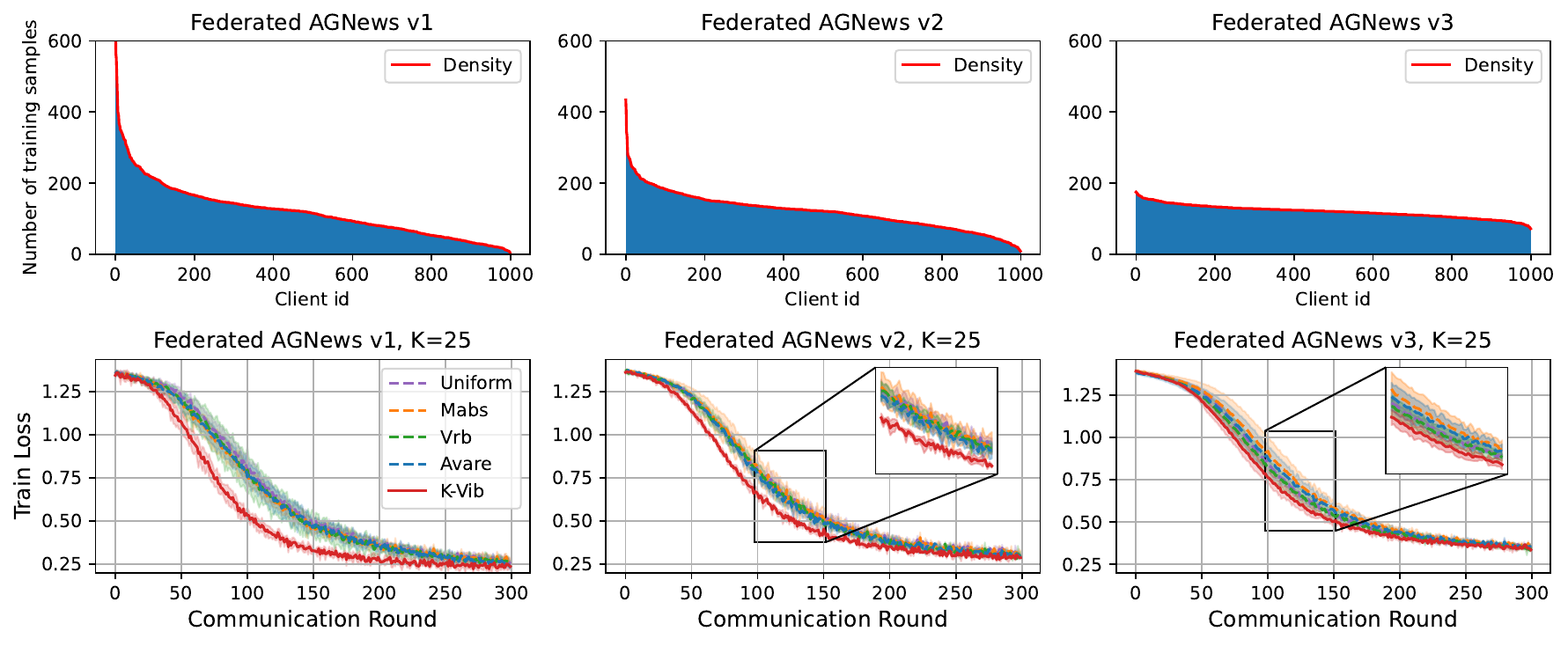}
\end{minipage}}
\hspace{0.5mm}
\subfigure[Experiments on CCNews dataset with Pythia-70M model.\label{fig:ccnews}]{
\begin{minipage}[t]{0.8\textwidth}
\includegraphics[width=\linewidth]{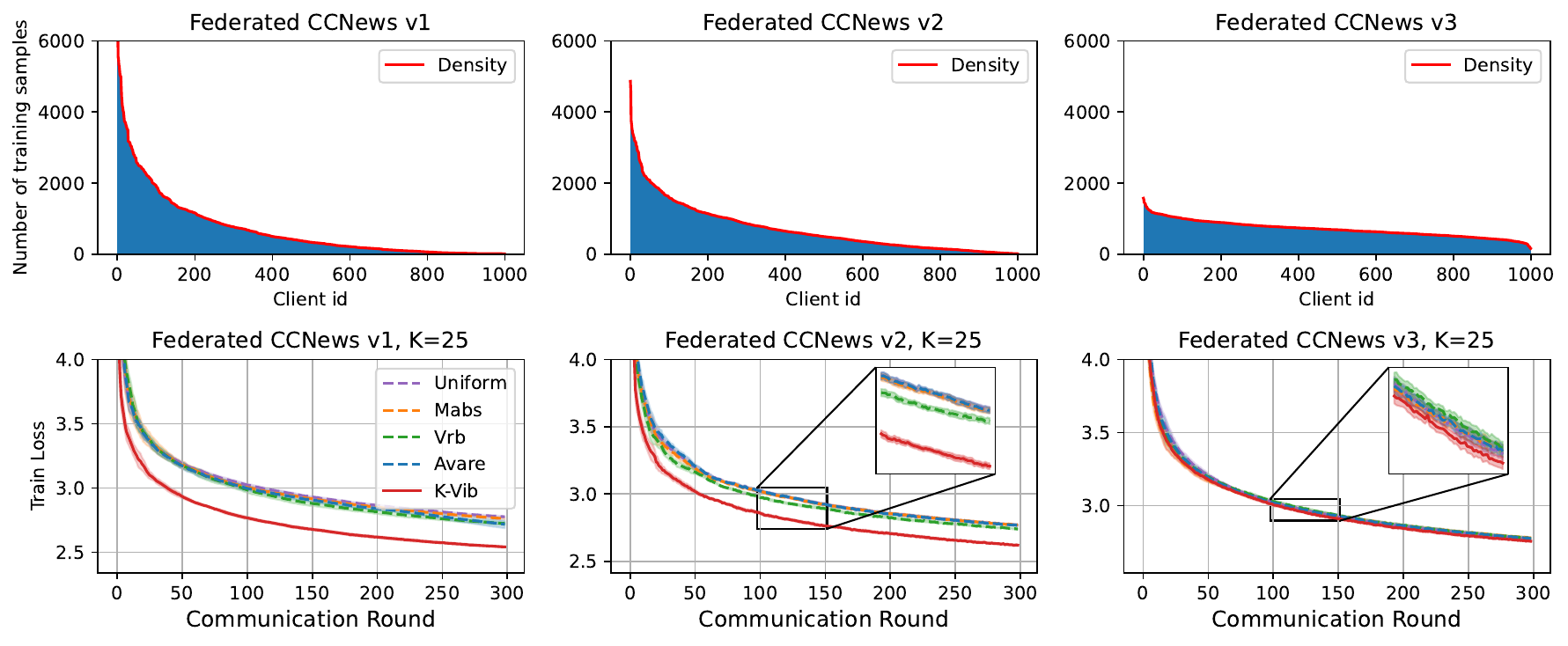}
\end{minipage}}
\hspace{0.5mm}
\caption{Federated text dataset experiments.}\label{fig:textdataset}
\end{figure*}

\subsection{CCNews and AGNews Datasets}

We evaluate the efficacy of the K-Vib sampler on two large sizes of models and datasets, including a fine-tuning task on AGNews~\citep{zhang2015character} and a pre-training task on CCNews~\citep{mackenzie2020cc}. AGNews is a text classification task with 119,999 train samples and 4 labels. And, CCNews is a text dataset that contains 708,241 articles. For models, we fine-tune a pretrained language model DistillBert~\citep{sanh2019distilbert} on the AGNews task. DistillBert is a language model with 67 Million parameters. And, we train \emph{from scratch} a GPT2 model called Pythia-70M~\citep{biderman2023pythia} (70 Millison parameters) on CCNews using the next token prediction loss.
For both tasks, we partition the datasets into $N=1, 000$ clients with three different levels of heavy long tails~\citep{charles2024towards}. Then, we set communication round $T=300$ and budget $K=25$. We set local learning rate $1e^{-4}$, batch size 16, and epoch 1 for the local SGD optimizer. 
We report the data distribution and the training loss for comparing the convergence benefits in Figure~\ref{fig:textdataset}. 

Analogous to FEMNIST experiments, we observe that K-Vib achieves $2\times$ faster convergence than baseline methods, while baseline methods only implement a marginal improvement compared to uniform sampling. And, the improvement is related to the degree of data variance across clients. Moreover, we surprisingly observed that baselines can be less efficient than uniform sampling in some cases. This is because ISP derives a loose upper bound, which can be less effective when estimating results of larger dimensions.
Therefore, our results prove that K-Vib can enhance real-world FL applications, even on large model training. 

\section{Discussion \& Conclusion}

\textbf{Extension \& Limitation.}\label{limitation} 
Our theoretical findings can be extended to general applications that involve estimating global results with partial information. Additionally, our extension of independent sampling can be applied to previous works employing random sampling. 
Besides, the global estimate variance in FL also comes from the data heterogeneity issues, which may incur unstable local feedback, breaking the Assumption~\ref{asp:local_convergence}.
This can be addressed with client clustering techniques~\citep{ghosh2020efficient, ma2022convergence, zeng2023stochastic}, analogous to previously cluster sampling works~\citep{fraboni2021clustered, song2023fast}. Besides, we can replace FedAvg with more stable FedAvg variants~\citep{sun2024understanding, zeng2023tackling}.

In conclusion, our study provides a thorough examination of FL frameworks utilizing unbiased client sampling techniques from an optimization standpoint. Our findings highlight the importance of designing unbiased sampling probabilities for the ISP to enhance the efficiency of FL. Building upon this insight, we further extend the range of adaptive sampling techniques and achieve substantial improvements. We are confident that our work will contribute to the advancement of client sampling techniques in FL, making them more applicable and beneficial in various practical scenarios.

\bibliography{main}
\bibliographystyle{tmlr}

\newpage
\appendix
\appendixpage
\startcontents[sections]
\printcontents[sections]{l}{1}{\setcounter{tocdepth}{5}}

\newpage
\section{Related Work}\label{sec:related}

Our paper contributes to the literature on the importance sampling in stochastic optimization, online convex optimization, and client sampling in FL.

\textbf{Importance Sampling.} Importance sampling is a non-uniform sampling technique widely used in stochastic optimization~\citep{katharopoulos2018not} and coordinate descent~\citep{richtarik2016optimal}. ~\cite{zhao2015stochastic, needell2014stochastic} connects the variance of the gradient estimates and the optimal sampling distribution is proportional to the per-sample gradient norm. The insights of sampling and optimization quality can be transferred into federated client sampling, as we summarised in the following two topics.

\textbf{Online Variance Reduction.} Our paper addresses the topic of online convex optimization for reducing variance. Variance reduction techniques are frequently used in conjunction with stochastic optimization algorithms~\citep{defazio2014saga, johnson2013accelerating} to enhance optimization performance. These same variance reduction techniques have also been proposed to quicken federated optimization~\citep{dinh2020federated, malinovsky2022variance}. On the other hand, online learning~\citep{shalev2012online} typically employs an exploration-exploitation paradigm to develop decision-making strategies that maximize profits. Although some studies have considered client sampling as a multi-armed bandit problem, they have only provided limited theoretical results~\citep{kim2020accurate, cho2020bandit, yang2021federated}. In an intriguing combination, certain studies~\citep{salehi2017stochastic, borsos2018online, borsos2019online} have formulated data sampling in stochastic optimization as an online learning problem. These methods were also applied to client sampling in FL by treating each client as a data sample in their original problem~\citep{zhao2021adaptive, el2020adaptive}.

\textbf{Client Sampling in FL.} 
Client sampling methods in FL fall under two categories: biased and unbiased methods. Unbiased sampling methods ensure objective consistency in FL by yielding the same expected value of results as global aggregation with the participation of all clients. In contrast, biased sampling methods converge to arbitrary sub-optimal outcomes based on the specific sampling strategies utilized. Additional discussion about biased and unbiased sampling methods is provided in Appendix~\ref{sec:diff}. Recent research has focused on exploring various client sampling strategies for both biased and unbiased methods. For instance, biased sampling methods involve sampling clients with probabilities proportional to their local dataset size~\citep{mcmahan2017communication}, selecting clients with a large update norm with higher probability~\citep{chen2020optimal}, choosing clients with higher losses~\citep{cho2020client}, and building a submodular maximization to approximate the full gradients~\citep{balakrishnan2022diverse}. Meanwhile, several studies~\citep{chen2020optimal, cho2020client} have proposed theoretically optimal sampling methods for FL utilizing the unbiased sampling framework, which requires all clients to upload local information before conducting sampling action. Moreover, cluster-based sampling~\citep{fraboni2021clustered, xu2021optimizing, shen2022fast} relies on additional clustering operations where the knowledge of utilizing client clustering can be transferred into other client sampling techniques.

\section{Useful Lemmas and Corollaries}\label{sec:lemmas_and_cors}
\subsection{Auxiliary Lemmas}\label{sec:aux_lema}

\begin{lemma}[Lemma 13, \citealp{borsos2018online}]\label{lemma:sum_a} For any sequence of numbers $c_1, \dots, c_T \in [0,1]$ the following holds:
$$
\sum_{t=1}^T \frac{c_t^4}{(c_{1:t}^2)^{3/2}} \leq 44,
$$
where $c_{1:t} = \sum_{\tau=1}^t c_\tau$.
\end{lemma}

\begin{lemma}\label{lemma:inq}
For an arbitrary set of $n$ vectors $\{\bs{a}_i\}_{i=1}^n, \bs{a}_i \in \mathbb{R}^d$,
\begin{equation}\label{eq:lemma1}
\left\|\sum_{i=1}^n \mathbf{a}_i\right\|^2 \leq n \sum_{i=1}^n\left\|\mathbf{a}_i\right\|^2.
\end{equation}
\end{lemma}

\begin{lemma}\label{axuliary:a}
    For random variables $z_1, \dots, z_n$, we have
\begin{equation}\label{eq:lemma2}
\mathbb{E}\left[\left\|z_1+\ldots+z_n\right\|^2\right] \leq n\mathbb{E}\left[\left\|z_1\right\|^2+\ldots+\left\|z_n\right\|^2\right].
\end{equation}
\end{lemma}

\begin{lemma}\label{axuliary:b}
    For independent, mean 0 random variables $z_1, \dots, z_n$, we have
\begin{equation}\label{eq:lemma3}
\mathbb{E}\left[\left\|z_1+\ldots+z_n\right\|^2\right]=\mathbb{E}\left[\left\|z_1\right\|^2+\ldots+\left\|z_n\right\|^2\right].
\end{equation}
\end{lemma}

\begin{lemma}[Tuning the stepsize~\citep{koloskova2020unified}]\label{lemma:constant_stepsize} For any parameters $r_0 \geq 0, b \geq 0, e \geq 0, d \geq 0$ there exists constant stepsize $\eta \leq \frac{1}{d}$ such that
$$
\Psi_T:=\frac{r_0}{\eta T}+b \eta+e \eta^2 \leq 2\left(\frac{b r_0}{T}\right)^{\frac{1}{2}}+2 e^{1 / 3}\left(\frac{r_0}{T}\right)^{\frac{2}{3}}+\frac{d r_0}{T}.       
$$
\begin{proof}
Choosing $\eta=\min \left\{\left(\frac{r_0}{bT}\right)^{\frac{1}{2}},\left(\frac{r_0}{eT}\right)^{\frac{1}{3}}, \frac{1}{d}\right\} \leq \frac{1}{d}$ we have three cases
\begin{itemize}
\item $\eta=\frac{1}{d}$ and is smaller than both $\left(\frac{r_0}{bT}\right)^{\frac{1}{2}}$ and $\left(\frac{r_0}{eT}\right)^{\frac{1}{3}}$, then
$$
\Psi_T \leq \frac{d r_0}{T}+\frac{b}{d}+\frac{e}{d^2} \leq\left(\frac{b r_0}{T}\right)^{\frac{1}{2}}+\frac{d r_0}{T}+e^{1 / 3}\left(\frac{r_0}{T}\right)^{\frac{2}{3}}
$$
\item  $\eta=\left(\frac{r_0}{bT}\right)^{\frac{1}{2}}<\left(\frac{r_0}{eT}\right)^{\frac{1}{3}}$, then
$$
\Psi_T \leq 2\left(\frac{r_0 b}{T}\right)^{\frac{1}{2}}+e\left(\frac{r_0}{bT}\right) \leq 2\left(\frac{r_0 b}{T}\right)^{\frac{1}{2}}+e^{\frac{1}{3}}\left(\frac{r_0}{T}\right)^{\frac{2}{3}}
$$
\item  The last case, $\eta=\left(\frac{r_0}{eT}\right)^{\frac{1}{3}}<\left(\frac{r_0}{bT}\right)^{\frac{1}{2}}$
$$
\Psi_T \leq 2 e^{\frac{1}{3}}\left(\frac{r_0}{T}\right)^{\frac{2}{3}}+b\left(\frac{r_0}{eT}\right)^{\frac{1}{3}} \leq 2 e^{\frac{1}{3}}\left(\frac{r_0}{T}\right)^{\frac{2}{3}}+\left(\frac{b r_0}{T}\right)^{\frac{1}{2}}
$$
\end{itemize}
\end{proof}
\end{lemma}

\begin{lemma}[Upper bound of local drift, \citealp{reddi2020adaptive}]\label{lemma:local_drift} Let Assumption~\ref{asp:unbiasedness}~\ref{asp:bounded-variance} hold. For all client $i\in[N]$ with arbitrary local iteration steps $r \in [R]$, the local drift can be bounded as follows,
$$
\mathbb{E}\left\|\bs{x}_i^{t,r}-\bs{x}^t\right\|^2 \leq 5R\eta_l^2 (\sigma_l^2 + 6R\sigma_g^2 + 6R\left\| \nabla f(\bs{x}^t) \right\|^2)
$$
\begin{proof}
For $r \in [R]$, we have
$$
\begin{aligned}
&\quad \mathbb{E}\left[\left\|\bs{g}_i^t\right\|^2\right] = \mathbb{E}\left\|\bs{x}_i^{t,r}-\bs{x}^t\right\|^2 = \mathbb{E}\left\|\bs{x}_i^{t, r-1}-\bs{x}^t-\eta_l \nabla F_i(\bs{x}_i^{t, r-1})\right\|^2 \\
= & \mathbb{E}\left\|\bs{x}_i^{t, r-1}-\bs{x}^t-\eta_l (\nabla F_i(\bs{x}_i^{t, r-1})\pm \nabla F_i(\bs{x}_i^{t, r-1}))\right\|^2 \\
= & \mathbb{E}\left\|\bs{x}_i^{t, r-1}-\bs{x}^t-\eta_l\nabla F_i(\bs{x}_i^{t, r-1}))\right\|^2 + \mathbb{E}\left\|\eta_l\left(\nabla F_i(\bs{x}_i^{t, r-1})-\nabla f_i\left(\bs{x}_i^{t, r-1}\right)\right)\right\|^2 \\
= & \mathbb{E}\left[\left\|\bs{x}_i^{t, r-1}-\bs{x}^t\right\|^2 -2 <\bs{x}_i^{t, r-1}-\bs{x}^t , \eta_l \nabla F_i(\bs{x}_i^{t, r-1})>+  \left\|\eta_l \nabla F_i(\bs{x}_i^{t, r-1})\right\|^2 \right] + \eta_l^2\sigma_l^2\\
= & \mathbb{E}\left[\left\|\bs{x}_i^{t, r-1}-\bs{x}^t\right\|^2 -2 <\frac{1}{\sqrt{2R-1}} (\bs{x}_i^{t, r-1}-\bs{x}^t) , \sqrt{2R-1}\eta_l \nabla F_i(\bs{x}_i^{t, r-1})>+  \left\|\eta_l \nabla F_i(\bs{x}_i^{t, r-1})\right\|^2 \right] + \eta_l^2\sigma_l^2 \\
\leq & \left(1+\frac{1}{2R-1}\right)\mathbb{E}\left[\left\|\bs{x}_i^{t, r-1}-\bs{x}^t\right\|^2\right] + 2R\mathbb{E}\left[\left\|\eta_l \nabla F_i(\bs{x}_i^{t, r-1})\right\|^2 \right] + \eta_l^2\sigma_l^2 \\
= & \left(1+\frac{1}{2R-1}\right)\mathbb{E}\left[\left\|\bs{x}_i^{t, r-1}-\bs{x}^t\right\|^2\right] + 2R\mathbb{E}\left[\left\|\eta_l\left(\nabla F_i(\bs{x}_{i}^{t, r-1})\pm \nabla f(\bs{x}^t) \pm \nabla f_i\left(\bs{x}^t\right) \right)\right\|^2 \right] + \eta_l^2\sigma_l^2 \\
\leq& \left(1+\frac{1}{2R-1}\right)\mathbb{E}\left\|\bs{x}_i^{t, r-1}-\bs{x}^t\right\|^2 + 6R\mathbb{E}\left[\left\|\eta_l\left(\nabla f_i\left(\bs{x}_i^{t, r-1}\right)-\nabla f_i\left(\bs{x}^t\right)\right)\right\|^2\right] + 6R\mathbb{E}\left[\left\|\eta_l\left(\nabla f_i\left(\bs{x}^t\right)\right)\right\|^2\right] + \eta_l^2\sigma_l^2 \\
\leq& \left(1 +\frac{1}{2R-1} + 6R\eta_l^2 L^2\right) \mathbb{E}\left\|\bs{x}_i^{t, r-1}-\bs{x}^t\right\|^2+\eta_l^2 (\sigma_l^2 + 6R \mathbb{E}\left\| \nabla f_i(\bs{x}^t) \right\|^2) \\
\end{aligned}
$$
Unrolling the recursion, we obtain
\begin{equation}
\begin{aligned}
\mathbb{E}\left\|\bs{x}_i^{t,r}-\bs{x}^t\right\|^2 \leq & \sum_{p=0}^{r-1} \left(1 + \frac{1}{2R-1} + 4R\eta_l^2 L^2\right)^p \eta_l^2 (\sigma_l^2 + 6R\mathbb{E}\left[\left\| \nabla f_i(\bs{x}^t) \right\|^2\right]) \\
\leq & (R-1) \left[\left(1 + \frac{1}{R-1}\right)^{R}-1\right]\eta_l^2 (\sigma_l^2 + 6R\mathbb{E}\left[\left\| \nabla f_i(\bs{x}^t) \right\|^2\right]) \\
\leq &  5R\eta_l^2 (\sigma_l^2 + 6R\sigma_g^2 + 6R\mathbb{E}\left\| \nabla f(\bs{x}^t) \right\|^2) \\
\leq & 5R\eta_l^2 \sigma_l^2 + 30R^2\eta_l^2\mathbb{E}\left\| \nabla f_i(\bs{x}^t) \right\|^2, \\
\end{aligned}
\end{equation}
where we use the fact that $(1+\frac{1}{R-1})^{R} \leq 5$ for $R>1$. 


\end{proof}
\end{lemma}

\subsection{Arbitrary Sampling}\label{sec:asampling}

In this section, we summarize the arbitrary sampling techniques and present key lemmas used in this paper. The arbitrary sampling is mainly used either for generating mini-batches of samples in stochastic algorithms~\citep{chambolle2018stochastic, richtarik2016optimal} or for coordinate descent optimization~\citep{qu2016coordinate}. In contrast, we explain the background in the context of federated optimization.

In detail, let $S$ denote a sampling, which is a random set-valued mapping with values in $2^{[N]}$, where $[N] := \{1,2,\dots, N\}$. An arbitrary sampling $S$ is generated by assigning probabilities to all $2^N$ subsets of $[N]$, which associates a \textit{probability matrix} $\mathbf{P} \in \mathbb{R}^{N\times N}$ defined by
$$
\mathbf{P}_{ij} := \text{Prob}(\{i,j\}\subseteq S).
$$
Thus, the \textit{probability vector} $p = (\bs{p}_1,\dots, p_N) \in \mathbb{R}^N$ is composed of the diagonal entries of $\mathbf{P}$, and $\bs{p}_i := \text{Prob}(i\in S)$. Furthermore, we say that $S$ is \textit{proper} if $\bs{p}_i > 0$ for all $i$. Thus, it incurs that
$$
    K := \mathbb{E}[|S|] = \text{Trace}(\mathbf{P}) = \sum_{i=1}^N \bs{p}_i.
$$

The definition of sampling can be naively transferred to the context of federated client sampling. We refer to $K$ as the expected number of sampled clients per round in FL. The following lemma plays a key role in our problem formulation and analysis. 

\begin{lemma}[Generalization of Lemma 1~\citet{horvath2019nonconvex}]\label{lemma:arbi}
    Let $\bs{a}_1, \bs{a}_2, \dots, \bs{a}_N$ be vectors in $\mathbb{R}^d$ and let $\bar{\bs{a}} = \sum_{i=1}^N \bs{\lambda}_i \bs{a}_i$ be their weighted average. Let $S$ be a proper sampling. Assume that there is $\bs{v} \in \mathbb{R}^N$ such that
    \begin{equation}\label{lemma:asum}
    \mathbf{P} - \bs{p}\bs{p}^t \preceq \textbf{Diag}(\bs{p}_1\bs{v}_1, \bs{p}_2\bs{v}_2, \dots, \bs{p}_N\bs{v}_N).
    \end{equation}
    Then, we have 
    \begin{equation}\label{lemma:inequal}
        \mathbb{E}_{S\sim p}\Bigg[ \bigg\| \sum_{i \in S} \frac{\bs{\lambda}_i \bs{a}_i}{\bs{p}_i} - \bar{\bs{a}} \bigg\|^2 \Bigg] \leq \sum_{i=1}^N \bs{\lambda}_i^2 \frac{\bs{v}_i}{\bs{p}_i} \|\bs{a}_i\|^2,
    \end{equation}
    where the expectation is taken over sampling $S$. Whenever~\eqref{lemma:asum} holds, it must be the case that
    $$
    \bs{v}_i \geq 1-\bs{p}_i.
    $$
    
    Moreover, The random sampling admits $\bs{v}_i = \frac{N-K}{N-1}$.The independent sampling admits $\bs{v}_i = 1-\bs{p}_i$ and makes~\eqref{lemma:inequal} hold as equality.

\end{lemma}

\begin{proof}
Let $\mathbb{I}_{i\in S} = 1$ if $i\in S$ and $\mathbb{I}_{i\in S}= 0$ otherwise. Similarly, let $\mathbb{I}_{i,j\in S} = 1$ if $i\in S$ and $\mathbb{I}_{i,j\in S}= 0$ otherwise. Note that $\mathbb{E}[\mathbb{I}_{i\in S}] = \bs{p}_i$ and $\mathbb{E}[\mathbb{I}_{i,j\in S}]=\mathbf{P}_{ij}$. Then, we compute the mean of estimates $\tilde{\bs{a}} := \sum_{i\in S}\frac{\bs{\lambda}_i \bs{a}_i}{\bs{p}_i}$:

$$
\mathbb{E}[\tilde{\bs{a}}] = \mathbb{E}[\sum_{i\in S}\frac{\bs{\lambda}_i \bs{a}_i}{\bs{p}_i}] = \mathbb{E}[\sum_{i=1}^N \frac{\bs{\lambda}_i \bs{a}_i}{\bs{p}_i} \mathbb{I}_{i\in S}] = \sum_{i=1}^N \frac{\bs{\lambda}_i \bs{a}_i}{\bs{p}_i} \mathbb{E}[\mathbb{I}_{i\in S}] = \sum_{i=1}^N \bs{\lambda}_i \bs{a}_i = \bar{\bs{a}}.
$$

Let $\textbf{A} = [\bs{\zeta}_1, \dots, \bs{\zeta}_N] \in \mathbb{R}^{d\times N}$, where $\bs{\zeta}_i = \frac{\bs{\lambda}_i \bs{a}_i}{\bs{p}_i}$, and let $\bs{e}$ be the vector of all ones in $\mathbb{R}^N$. We now write the variance of $\tilde{\bs{a}}$ in a form that will be convenient to establish a bound:

\begin{equation}\label{eq:arbiv}
\begin{aligned} 
\mathbb{E}[\|\tilde{\bs{a}}-\mathbb{E}[\tilde{\bs{a}}]\|^2] & =\mathbb{E}[\|\tilde{\bs{a}}\|^2]-\|\mathbb{E}[\tilde{\bs{a}}]\|^2 \\ 
& =\mathbb{E}[\|\sum_{i \in S} \frac{\bs{\lambda} \bs{a}_i}{\bs{p}_i}\|^2]-\|\bar{\bs{a}}\|^2 \\ & =\mathbb{E}\left[\sum_{i, j} \frac{\bs{\lambda}_i \bs{a}_i^{\top}}{\bs{p}_i} \frac{\bs{\lambda}_j \bs{a}_j}{\bs{p}_j} \mathbb{I}_{i, j \in S}\right]-\|\bar{\bs{a}}\|^2 \\ 
& =\sum_{i, j} \bs{p}_{i j} \frac{\bs{\lambda}_i \bs{a}_i^{\top}}{\bs{p}_i} \frac{\bs{\lambda}_j \bs{a}_j}{\bs{p}_j}-\sum_{i, j} \bs{\lambda}_i \bs{\lambda}_j \bs{a}_i^{\top}\bs{a}_j \\ 
& =\sum_{i, j} \left(\bs{p}_{i j}-\bs{p}_i \bs{p}_j\right) \bs{\zeta}_i^{\top} \bs{\zeta}_j \\ 
& = \bs{e}^{\top}\left(\left(\mathbf{P}-\bs{p} \bs{p}^{\top}\right) \circ \mathbf{A}^{\top} \mathbf{A}\right) \bs{e} .
\end{aligned}
\end{equation}

Since by assumption we have $\mathbf{P}-\bs{p} \bs{p}^{\top} \preceq \mathbf{D i a g}(\bs{p} \circ \bs{v})$, we can further bound
\begin{equation}\label{eq:independent}
\bs{e}^{\top}\left(\left(\mathbf{P}-\bs{p} \bs{p}^{\top}\right) \circ \mathbf{A}^{\top} \mathbf{A}\right) \bs{e} \leq \bs{e}^{\top}\left(\mathbf{D i a g}(\bs{p} \circ \bs{v}) \circ \mathbf{A}^{\top} \mathbf{A}\right) \bs{e}=\sum_{i=1}^n \bs{p}_i \bs{v}_i \left\|\bs{\zeta}_i\right\|^2.
\end{equation}

To obtain \eqref{lemma:inequal}, it remains to combine \eqref{eq:independent} with \eqref{eq:arbiv}. Since $\mathbf{P}-\bs{p} \bs{p}^{\top}$ is positive semi-definite~\citep{richtarik2016parallel}, we can bound $\mathbf{P}-\bs{p} \bs{p}^{\top} \preceq N \mathbf{D i a g}(\mathbf{P}-\bs{p} \bs{p}^{\top}) = \mathbf{D i a g}(\bs{p} \circ \bs{v})$, where $\bs{v}_i = N(1-\bs{p}_i)$. 

Overall, arbitrary sampling that associates with a probability matrix $\mathbf{P}$ will determine the value of $\bs{v}$. As a result, we summarize independent sampling and random sampling as follows,

\begin{itemize}
    \item Consider now the independent sampling,
    $$
    \mathbf{P}-\bs{p} \bs{p}^{\top}=\left[\begin{array}{cccc}
    \bs{p}_1\left(1-\bs{p}_1\right) & 0 & \cdots & 0 \\
    0 & \bs{p}_2\left(1-\bs{p}_2\right) & \cdots & 0 \\
    \vdots & \vdots & \ddots & \vdots \\
    0 & 0 & \cdots & \bs{p}_n\left(1-\bs{p}_n\right)
    \end{array}\right]=\mathbf{Diag}\left(\bs{p}_1 \bs{v}_1, \ldots, \bs{p}_n \bs{v}_n\right),
    $$
    where $\bs{v}_i = 1 - \bs{p}_i$. Therefore, independent sampling always minimizes \eqref{lemma:inequal}, making it hold as equality.

    \item Consider the random sampling,
    $$
    \mathbf{P}-\bs{p} \bs{p}^{\top}=\left[\begin{array}{cccc}
    \frac{K}{N}-\frac{K^2}{N^2} & \frac{K(K-1)}{N(N-1)} & \cdots & \frac{K(K-1)}{N(N-1)} \\
    \frac{K(K-1)}{N(N-1)} & \frac{K}{N} & \cdots & \frac{K(K-1)}{N(N-1)} \\
    \vdots & \vdots & \ddots & \vdots \\
    \frac{K(K-1)}{N(N-1)} & \frac{K(K-1)}{N(N-1)} & \cdots & \frac{K}{N}
    \end{array}\right].
    $$
    As shown in \citep{horvath2019nonconvex}, the standard random sampling admits $\bs{v}_i = \frac{N-K}{N-1}$ for \eqref{lemma:inequal}.
\end{itemize}
\end{proof}
\textit{\textbf{Conclusion.}} Given probabilities $\bs{p}$ that defines all samplings $S$ satisfying $\bs{p}_i = \text{Prob}(i\in S)$, it turns out that the independent sampling (i.e., $\mathbf{P}_{ij} = \text{Prob}(i,j\in S) = \text{Prob}(i\in S)\text{Prob}(j\in S)=\bs{p}_i\bs{p}_j$) minimizes the upper bound in~\eqref{lemma:inequal}. Therefore, depending on the sampling distribution and method, we can rewrite the~\eqref{lemma:inequal} as follows:

\begin{equation}
    \mathbb{V}(S) =  \mathbb{E}_{S \sim \bs{p}}[\|\sum_{i\in S}\frac{\bs{\lambda}_i \bs{a}_i}{\bs{a}_i} - \bar{\bs{a}}\|^2]  = \underbrace{\sum_{i=1}^N (1-\bs{p}_i) \frac{\bs{\lambda}_i^2\|\bs{a}_i\|^2}{\bs{p}_i}}_{\text{Independent sampling procedure}} \leq \underbrace{\frac{N-K}{N-1}\sum_{i=1}^N \frac{\bs{\lambda}_i^2\|\bs{a}_i\|^2}{\bs{p}_i}}_{\text{Random sampling procedure}}.
\end{equation}



\subsection{Proof of Solution to Independent Sampling with Minimal Probability}\label{sec:keylemmas}

In this section, we present lemmas and their proofs for our theoretical analyses. Our methodology of independent sampling especially guarantees a minimum probability of clients in comparison with Lemma~\ref{lemma:optimal}. Our proof involves a general constraint, which covers Lemma~\ref{lemma:optimal}. Then, we provide several Corollaries~\ref{cor:min}~\ref{cor:pmin-bound} for our analysis in the next section.

\begin{lemma}\label{obj:restrict}
Let $ 0 < \bs{a}_1 \leq \bs{a}_2 \leq \dots \leq \bs{a}_N$ and $0 < K \leq N$. We consider the following optimization objective with a restricted probability space $\Delta = \{\bs{p} \in \mathbb{R}^N | p_{\text{min}} \leq \bs{p}_i \leq 1, \sum_{i=1}^N \bs{p}_i = K, \forall i \in [N]\}$ where $p_{\text{min}} \leq K/N$,

\begin{equation}\label{obj:restricted}
\begin{aligned}
\text{minimize}_{\bs{p} \in \Delta} \;& \Omega(\bs{p}) = \sum_{i=1}^N \frac{\bs{a}_i^2}{\bs{p}_i}\\
\text{subject to}\; & \sum_{i=1}^N \bs{p}_i = K, \\
     & p_{\text{min}} \leq \bs{p}_i \leq 1, \; i = 1,2,\dots,N.
\end{aligned}
\end{equation}
\end{lemma}

\begin{proof}
We formulate the Lagrangian:
\begin{equation}
\mathcal{L}(p, y, \alpha_1, \dots, \alpha_N, \beta_1, \dots, \beta_N) = \sum_{i=1}^N \frac{\bs{a}_i^2}{\bs{p}_i} + y \cdot \Big( \sum_{i=1}^N \bs{p}_i - K\Big) + \sum_{i=1}^N \alpha_i (p_{\text{min}} - \bs{p}_i ) + \sum_{i=1}^N \beta_i (\bs{p}_i - 1).
\end{equation}

The constraints are linear and KKT conditions hold. Hence, we have,
\begin{equation}\label{obj:optimalp}
    \bs{p}_i = \sqrt{\frac{\bs{a}_i^2}{y-\alpha_i+\beta_i}} = \begin{cases}
        1, & \text{if}\; \sqrt{y} \leq \bs{a}_i.\\
        \sqrt{\frac{\bs{a}_i^2}{y}}, & \text{if}\; \sqrt{y}\cdot p_{\text{min}} < \bs{a}_i < \sqrt{y}, \\
        p_{\text{min}}, & \text{if}\; \bs{a}_i \leq \sqrt{y}\cdot p_{\text{min}}. \\
    \end{cases}
\end{equation}

Then, we analyze the value of $y$. Letting $l_1 = \big|\{i |  \bs{a}_i \leq \sqrt{y}\cdot p_{\text{min}} \}\big| $, $l_2 = l_1+|\{\sqrt{y}\cdot p_{\text{min}} < \bs{a}_i < \sqrt{y}\}|$, $N-l_2 = \big|\{i |  \sqrt{y} \leq \bs{a}_i \}\big|$, and using $\sum_{i=1}^N \bs{p}_i = K$ implies,
$$
\sum_{i=1}^N \bs{p}_i = \sum_{i \leq l_1} \bs{p}_i + \sum_{l_1 < i < l_2} \bs{p}_i + \sum_{i \geq l_2} \bs{p}_i = l_1\cdot p_{\min} +  \sum_{l_1 < i < l_2} \sqrt{\frac{\bs{a}_i^2}{y}} + N  - l_2 = K.
$$

Arrange the formula, we get
\begin{equation}\label{eq:optimaly}
    \sqrt{y} = \frac{\sum_{l_1 < i < l_2} \bs{a}_i}{K - N + l_2 - l_1 \cdot p_{\text{min}}}.
\end{equation}

Moreover, we can plug the results into the objective to get the optimal result:
\begin{equation}\label{eq:pmin-result}
\begin{aligned}
\sum_{i=1}^N \frac{\bs{a}_i^2}{\bs{p}_i} & = \sum_{i \leq l_1} \frac{\bs{a}_i^2}{\bs{p}_i} + \sum_{l_1 < i < l_2} \frac{\bs{a}_i^2}{\bs{p}_i} + \sum_{i \geq N-l_2} \frac{\bs{a}_i^2}{\bs{p}_i} \\ 
& = \frac{\sum_{i \leq l_1} \bs{a}_i^2}{p_{\text{min}}} + \sqrt{y}(\sum_{l_1 < i < l_2} \bs{a}_i) + \sum_{i \geq N-l_2} \bs{a}_i^2 \\
& = \frac{\sum_{i \leq l_1} \bs{a}_i^2}{p_{\text{min}}} + \frac{(\sum_{l_1 < i < l_2} \bs{a}_i)^2}{K-N+(l_2-l_1\cdot p_{\text{min}})} + \sum_{i \geq N-l_2} \bs{a}_i^2, \\
\end{aligned}
\end{equation}

where the $1 \leq l_1 \leq l_2 \leq N$, which satisfies that $\forall i \in (l_1, l_2)$,
$$
p_{\text{min}} \cdot \frac{\sum_{l_1 < i < l_2} \bs{a}_i}{K - N + l_2 - l_1 \cdot p_{\text{min}}} < \bs{a}_i < \frac{\sum_{l_1 < i < l_2} \bs{a}_i}{K - N + l_2 - l_1 \cdot p_{\text{min}}}.
$$

In short, we note that if let $p_{\text{min}} = 0, l_1 = 0$, the Lemma~\ref{lemma:optimal} is proved as a special case of \eqref{eq:pmin-result}. Besides, we provide further Corollary~\ref{cor:min} and \ref{cor:pmin-bound} as preliminaries for further analysis.





\begin{corollary}\label{cor:min}
With $K\cdot a_N \leq \sum_{i=1}^N \bs{a}_i$ and $p_{\text{min}}=0$, we have $l_1=0, l_2=N$ for \eqref{eq:pmin-result} and induce
$$
\arg \min \Omega(\bs{p}^*) = \frac{(\sum_{i=1}^N \bs{a}_i)^2}{K}.
$$
\end{corollary}



\begin{corollary}\label{cor:pmin-bound}
With $K\cdot a_N \leq \sum_{i=1}^N \bs{a}_i$ and $p_{\text{min}}>0$, we have $l_2 = N$ and $l_1$ is the largest integer that satisfies $0 < (K-l_1\cdot p_{\text{min}})\frac{a_{l_1}}{\sum_{i=l_1}^{N} \bs{a}_i} < p_{\text{min}}$. The optimal value of \eqref{eq:pmin-result} becomes
$$
\begin{aligned}
\sum_{i=1}^N \frac{\bs{a}_i^2}{\bs{p}_i} & = \frac{\sum_{i \leq l_1} \bs{a}_i^2}{p_{\text{min}}} + \sqrt{y}(\sum_{l_1 < i \leq N} \bs{a}_i)  & \text{$\triangleright$ Eq.~\eqref{eq:pmin-result}, def. in line 2}\\
& = \frac{\sum_{i \leq l_1} \bs{a}_i^2}{p_{\text{min}}} + y(K-l_1 p_{\text{min}}) & \text{$\triangleright$ Eq.~\ref{eq:optimaly}, replacing $\sum_{l_1 < i \leq N} \bs{a}_i$} \\
& \leq l_1 y p_{\text{min}} + y(K-l_1 p_{\text{min}}) & \text{$\triangleright$ Eq.~\ref{obj:optimalp}, $\bs{a}_i \leq \sqrt{y}\cdot p_{\text{min}}$} \\
& = \frac{(\sum_{i={l_1}}^{N} \bs{a}_i )^2}{(K-l_1 p_{\text{min}})^2} \cdot K \leq \frac{K(\sum_{i={l_1}}^{N} \bs{a}_i )^2}{(K-Np_{\text{min}})^2} & \\
& \leq \frac{K(\sum_{i={1}}^{N} \bs{a}_i )^2}{(K-Np_{\text{min}})^2}. &
\end{aligned}
$$
\end{corollary}
\end{proof}

\section{Convergence Analyses}\label{sec:convergence_analysis}

\subsection{Analysis on Sampling}\label{sec:conv_pre}
We start our convergence analysis with a clarification of the concepts of optimal independent sampling. Considering an Oracle always outputs the optimal probabilities $\bs{p}^*$, we define
$$
\delta_*^t := \mathbb{E}\left[\left\|\sum_{i\in S^*} \frac{\bs{\lambda}_i \bs{g}_i^t}{\bs{p}_i^*} - \sum_{i=1}^N \bs{\lambda}_i \bs{g}_i^t\right\|^2\right] = \mathbb{E}\left[ \sum_{i=1}^N \frac{1-\bs{p}_i^*}{\bs{p}_i^*} \|\tilde{\bs{g}}_i^t\|^2\right],
$$
where we have $\|\tilde{\bs{g}}_i^t\|^2 = \|\bs{\lambda}_i \bs{g}_i^t\|^2$.  Then, we plug the optimal probability in \eqref{eq:optimal_p} into the above equation to obtain
$$
\delta_*^t = \mathbb{E}\left[ \sum_{i=1}^N \frac{1-\bs{p}_i^*}{\bs{p}_i^*} \|\tilde{\bs{g}}_i^t\|^2\right] = \mathbb{E}\left[\frac{1}{K-(N-l)}\left(\sum_{i=1}^l \|\tilde{\bs{g}}_i^t\|\right)^2 - \sum_{i=1}^l \|\tilde{\bs{g}}_i^t\|^2\right].
$$
Using the fact that $K \|\tilde{\bs{g}}_N^t\| \leq \sum_{i=1}^N \|\tilde{\bs{g}}_i^t\|$, we have
$$
\begin{aligned}
\delta_*^t & \leq \mathbb{E}\left[\frac{1}{K}\left(\sum_{i=1}^N \|\tilde{\bs{g}}_i^t\|\right)^2 - \sum_{i=1}^N \|\tilde{\bs{g}}_i^t\|^2\right] \\
& = \mathbb{E}\left[\frac{1}{K}\left(\sum_{i=1}^N \|\tilde{\bs{g}}_i^t\|\right)^2 \left(1- K \frac{\sum_{i=1}^N \|\tilde{\bs{g}}_i^t\|^2}{\left(\sum_{i=1}^N \|\tilde{\bs{g}}_i^t\|\right)^2}\right)\right] \\
& \leq \frac{N-K}{NK}\mathbb{E}\left[ \left(\sum_{i=1}^N \|\tilde{\bs{g}}_i^t\|\right)^2 \right].
\end{aligned}
$$

To clarify the improvement of utilizing the sampling procedure, we provide two baseline analyses respecting independent sampling and random sampling. For an uniform independent sampling $S^t \sim \mathbb{U}(\bs{p}_i = \frac{K}{N})$ , we have
\begin{equation}\label{eq:delta_uniform}
\delta^t_{\mathbb{U}} := \mathbb{E}\left[\left\|\sum_{i\in S^t} \frac{\bs{\lambda}_i}{\bs{p}_i}\bs{g}_i^t - \sum_{i=1}^N \bs{\lambda}_i \bs{g}_i^t \right\|^2\right] = \mathbb{E}\left[\sum_{i=1}^N \frac{1-\frac{K}{N}}{\frac{K}{N}} \|\tilde{\bs{g}}_i^t\|^2\right] = \frac{N-K}{K} \mathbb{E}\left[\sum_{i=1}^N \|\tilde{\bs{g}}_i^t\|^2\right].
\end{equation}


Straightforwardly, we can prove that $\delta^t/\delta_{\mathbb{U}}<0$, indicating that independent sampling creates better estimates than random sampling.

\begin{definition}[The optimal factor] Given an iteration sequence of global model $\{\bs{x}^1, \dots, \bs{x}^t\}$, under the constraints of communication budget $K$ and local updates statues $\{\bs{g}_i^t\}_{i\in[N]}, t \in [T]$, we define the improvement factor of applying optimal client sampling $S^t_*\sim \bs{p}^*$ comparing uniform sampling $U^t \sim \mathbb{U}$ as:
$$
\alpha^t_* = \frac{\mathbb{E}\left[\left\|\sum_{i\in S^t_*} \frac{\bs{\lambda}_i}{\bs{p}_i^*} \bs{g}_i^t - \sum_{i=1}^N \bs{\lambda}_i \bs{g}_i^t\right\|^2\right]}{\mathbb{E}\left[\left\|\sum_{i\in U^t} \frac{\bs{\lambda}_i}{\bs{p}_i}\bs{g}_i^t - \sum_{i=1}^N \bs{\lambda}_i \bs{g}_i^t \right\|^2\right]},
$$
and optimal $\bs{p}^*$ is computed via \eqref{eq:optimal_p} with $\{\bs{g}^t_i\}_{i\in[N]}$. Moreover, we can know
\begin{equation}\label{eq:optimal_ratio}
\begin{aligned}
\alpha_*^t & := \frac{\delta_*^t}{\delta_{\mathbb{U}}} = \frac{\mathbb{E}\left[\left\|\sum_{i\in S^*} \frac{\bs{\lambda}_i}{\bs{p}_i^*} \bs{g}_i^t - \sum_{i=1}^N \bs{\lambda}_i \bs{g}_i^t\right\|^2\right]}{\mathbb{E}\left[\left\|\sum_{i\in S^t} \frac{\bs{\lambda}_i}{\bs{p}_i}\bs{g}_i^t - \sum_{i=1}^N \bs{\lambda}_i \bs{g}_i^t \right\|^2\right]} \\
& \leq \frac{K\mathbb{E}\left[ \left(\sum_{i=1}^N \|\tilde{\bs{g}}_i^t\|\right)^2 \right]}{NK \mathbb{E}\left[\sum_{i=1}^N \|\tilde{\bs{g}}_i^t\|^2\right]} < \frac{\mathbb{E}\left[ \left(\sum_{i=1}^N \|\tilde{\bs{g}}_i^t\|\right)^2 \right]}{N \mathbb{E}\left[\sum_{i=1}^N \|\tilde{\bs{g}}_i^t\|^2\right]} \leq 1.
\end{aligned}
\end{equation}
\end{definition}

\subsection{Non-convex Analyses}\label{sec:main_analysis}

Now we are ready to give our convergence analysis in detail.
\begin{proof}
We recall the updating rule during round $t$ as: 
$$
\bs{x}^{t+1} = \bs{x}^t - \eta_g \sum_{i\in S^t} \frac{\bs{\lambda}_i \bs{g}_i^t}{\bs{p}_i^t} := \bs{x}^t - \eta_g \bs{d}^t, \text{where} \; \bs{g}_i^t = \bs{x}^t - \bs{x}^{t, R}_i = \eta_l \sum_{r=1}^{R} \nabla F_i(\bs{x}^{t, r-1}_i). 
$$

Without loss of generality, we rewrite the global descent rule as:
$$
\bs{x}^{t+1} = \bs{x}^t - \frac{\eta}{R} \sum_{i\in S^t} \frac{\bs{\lambda}_i}{\bs{p}_i^t} \tilde{\boldsymbol{g}}_i^t := \bs{x}^t - \eta \tilde{\boldsymbol{d}}^t,
$$
where $\eta = R \eta_l \eta_g$, $\tilde{\boldsymbol{g}}_i^t = \sum_{r=1}^{R} \nabla F_i(\bs{x}^{t, r-1}_i)$. Therefore, we know $\mathbb{E}_{S^t}[\tilde{\boldsymbol{d}}^t] = \frac{1}{R}\sum_{i=1}^N \bs{\lambda}_i \tilde{\boldsymbol{g}}_i^t$ and $\mathbb{E}\|\boldsymbol{d}^t\| = \frac{\eta^2}{R^2\eta_l^2}\mathbb{E}\|\tilde{\boldsymbol{d}}^t\|$. Moreover, we denote $W = \max \{\bs{\lambda}_i\}_{i\in[N]}$.

\textbf{Descent lemma.} Using the smoothness of $f$ and taking expectations conditioned on $x$ and over the sampling $S^t$, we have
\begin{equation}\label{eq:descent}
\begin{aligned}
\mathbb{E}\left[f(\bs{x}^{t+1})\right] & = \mathbb{E}\left[f(\bs{x}^t - \eta \tilde{\boldsymbol{d}}^t)\right] \leq \mathbb{E}[f(\bs{x}^t)] - \eta \mathbb{E}[\left \langle \nabla f(\bs{x}^t), \mathbb{E}_{S^t}[\tilde{\boldsymbol{d}}^t] \right \rangle] + \frac{L}{2}\eta^2 \mathbb{E}\left[\|\tilde{\boldsymbol{d}}^t\|^2\right] \\
& \leq \mathbb{E}[f(\bs{x}^t)] - \eta \mathbb{E}\|\nabla f(\bs{x}^t)\|^2 + \eta \mathbb{E}[\left \langle \nabla f(\bs{x}^t), \nabla f(\bs{x}^t) - \mathbb{E}_{S^t}[\tilde{\boldsymbol{d}}^t] \right \rangle] + \frac{L}{2}\eta^2 \mathbb{E}\left[\|\tilde{\boldsymbol{d}}^t\|^2\right] \\
& \leq f(\bs{x}^t) - \frac{\eta}{2} \|\nabla f(\bs{x}^t)\|^2 +  \frac{\eta}{2}\underbrace{\mathbb{E}\left[\| \nabla f(\bs{x}^t) - \mathbb{E}_{S^t}[\tilde{\boldsymbol{d}}^t] \|^2\right]}_{T_1} + \frac{L}{2}\eta^2 \underbrace{\mathbb{E}\left[\|\tilde{\boldsymbol{d}}^t\|^2\right]}_{T_2}, \\
\end{aligned}
\end{equation}
where the last inequality follows since $\langle a,b \rangle \leq \frac{1}{2}\|a\|^2 + \frac{1}{2}\|b\|^2, \forall a,b \in \mathbb{R}^d$. 

\textbf{Bounding $T_1$.} 
We first investigate the expectation gap between global first-order gradient and utilized global estimates,
$$
\begin{aligned}
& \quad \mathbb{E}\left[\| \nabla f(\bs{x}^t) - \mathbb{E}_{S^t}[\tilde{\boldsymbol{d}}^t] \|^2\right] = \mathbb{E}\left[\left\| \sum_{i=1}^N \bs{\lambda}_i \nabla f_i(\bs{x}^t) - \frac{1}{R}\sum_{i=1}^N \bs{\lambda}_i \tilde{\bs{g}}_i^t \right\|^2\right] \\
& = \mathbb{E}\left[\left\| \sum_{i=1}^N \bs{\lambda}_i \left(\nabla f_i(\bs{x}^t) - \frac{1}{R} \sum_{r=1}^R \nabla F_i(\bs{x}^{t, r-1}_i) \right) \right\|^2\right] \\
& = \mathbb{E}\left[\left\| \sum_{i=1}^N \bs{\lambda}_i \frac{1}{R} \sum_{r=1}^R\left(\nabla f_i(\bs{x}^t) -\nabla F_i(\bs{x}^{t, r-1}_i)+\nabla f_i(\bs{x}^{t, r-1}_i)- \nabla f_i(\bs{x}^{t, r-1}_i) \right) \right\|^2\right] \\
& \leq 2\mathbb{E}\left[\left\| \sum_{i=1}^N \bs{\lambda}_i \frac{1}{R} \sum_{r=1}^R\left(\nabla f_i(\bs{x}^t) -\nabla f_i(\bs{x}^{t, r-1}_i)\right) \right\|^2\right] +2\mathbb{E}\left[\left\| \sum_{i=1}^N \bs{\lambda}_i \frac{1}{R} \sum_{r=1}^R\left(\nabla f_i(\bs{x}^{t, r-1}_i)- \nabla F_i(\bs{x}^{t, r-1}_i) \right) \right\|^2\right] \\
& \leq 2 \mathbb{E}\left[\left\| \sum_{i=1}^N \bs{\lambda}_i \frac{1}{R} \sum_{r=1}^R\left(\nabla f_i(\bs{x}^t) -\nabla f_i(\bs{x}^{t, r-1}_i)\right) \right\|^2\right] + 2\frac{\sigma_l^2}{R} \\
& \leq 2L^2 \sum_{i=1}^N \bs{\lambda}_i \frac{1}{R}\sum_{r=1}^R \mathbb{E}\left\|\bs{x}^t - \bs{x}^{t, r-1}_i \right\|^2 + 2\frac{\sigma_l^2}{R} \\
& \leq 2L^2 \sum_{i=1}^N \bs{\lambda}_i (5R\eta_l^2\sigma_l^2 + 30R^2\eta_l^2\mathbb{E}\left\| \nabla f_i(\bs{x}^t) \right\|^2) + 2 \frac{\sigma_l^2}{R} \quad\quad\quad \triangleright \text{use Lemma~\ref{lemma:local_drift}} \\
& \leq 10RL^2 \eta_l^2 \sigma_l^2 + 60R^2L^2\eta_l^2 \sum_{i=1}^N \bs{\lambda}_i \mathbb{E}\left\| \nabla f_i(\bs{x}^t) \right\|^2 + 2\frac{\sigma_l^2}{R} \\
& \leq 10RL^2 \eta_l^2 \sigma_l^2 + 60R^2L^2\eta_l^2 (\|\nabla f(\bs{x}^t)\|^2 + \sigma_g^2) + 2\frac{\sigma_l^2}{R} \\
& \leq 60R^2L^2\eta_l^2 \|\nabla f(\bs{x}^t)\|^2 + (10RL^2 \eta_l^2 + \frac{2}{R})\sigma_l^2 + 60R^2L^2\eta_l^2 \sigma_g^2,
\end{aligned}
$$
where we use the fact by Assumption~\ref{asp:bounded-variance} that
$\sum_{i=1}^N \bs{\lambda}_i \|\nabla f_i(\bs{x}^t)\|^2 \leq \|\nabla f(\bs{x}^t)\|^2 + \sigma_g^2$. Without loss of generality, we consider $10RL^2 \eta_l^2$ to dominate the coefficients of the second term above with $R \geq 2$. Therefore, we have 
\begin{equation}
\begin{aligned}
\mathbb{E}\left[\| \nabla f(\bs{x}^t) - \mathbb{E}_{S^t}[\tilde{\boldsymbol{d}}^t] \|^2\right] \leq 60R^2L^2\eta_l^2 \|\nabla f(\bs{x}^t)\|^2 + 10RL^2 \eta_l^2\sigma_l^2 + 60R^2L^2\eta_l^2 \sigma_g^2.
\end{aligned}
\end{equation}

If we further restrict $\eta \leq \frac{1}{8L}$, then for any $\eta_g \geq 1$ which induces $\eta_l \leq \frac{1}{8RL}$, we have
\begin{equation}\label{eq:t1}
T_1 = \mathbb{E}\left[\| \nabla f(\bs{x}^t) - \mathbb{E}_{S^t}[\tilde{\boldsymbol{d}}^t] \|^2\right] \leq \frac{15}{16}\|\nabla f(\bs{x}^t)\|^2 + 10L^2 \sigma^2 \frac{\eta^2}{\eta_g^2},
\end{equation}
where we define $\sigma^2 = \frac{\sigma_l^2}{R} + 6\sigma_g^2$.

\textbf{Bounding $T_2$.} 
Now, we need to bound estimates:
\begin{equation}\label{eq:estimates}
\begin{aligned}
\mathbb{E}\left[\|\bs{d}^t\|^2\right] & \leq \mathbb{E}\left[\left\|\sum_{i\in S^t} \frac{\bs{\lambda}_i \bs{g}_i^t}{\bs{p}_i^t} - \sum_{i=1}^N \bs{\lambda}_i \bs{g}_i^t\right\|^2 + \left\|\sum_{i=1}^N \bs{\lambda}_i \bs{g}_i^t\right\|^2\right] \\
& \leq \underbrace{\mathbb{E}\left[\left\|\sum_{i\in S^t} \frac{\bs{\lambda}_i \bs{g}_i^t}{\bs{p}_i^t} - \sum_{i\in S^*} \frac{\bs{\lambda}_i \bs{g}_i^t}{\bs{p}_i^*}\right\|^2\right]}_{Q(S^t)} + \underbrace{\mathbb{E}\left[\left\|\sum_{i\in S^*} \frac{\bs{\lambda}_i \bs{g}_i^t}{\bs{p}_i^*} - \sum_{i=1}^N \bs{\lambda}_i \bs{g}_i^t\right\|^2\right] + \mathbb{E}\left[\left\|\sum_{i=1}^N \bs{\lambda}_i \bs{g}_i^t\right\|^2\right]}_{(A)}. \\
\end{aligned}
\end{equation}
Here, the $Q(S^t)$ indicates the discrepancy between applied sampling and optimal sampling. The term $(A)$ indicates the intrinsic gap for the optimal sampling to approach its targets and the quality of the targets for optimization. Using \eqref{eq:delta_uniform} and \eqref{eq:optimal_ratio}, we have 
$$
\begin{aligned}
(A) & = \mathbb{E}\left[\left\|\sum_{i\in S^*} \frac{\bs{\lambda}_i \bs{g}_i^t}{\bs{p}_i^*} - \sum_{i=1}^N \bs{\lambda}_i \bs{g}_i^t\right\|^2\right] + \mathbb{E}\left\|\sum_{i=1}^N \bs{\lambda}_i \bs{g}_i^t\right\|^2 \\
& \leq \alpha_*^t \frac{N-K}{K} \mathbb{E}\left[ \sum_{i=1}^N \bs{\lambda}_i^2 \left\|\bs{g}_i^t\right\|^2\right] + \mathbb{E}\left\|\sum_{i=1}^N \bs{\lambda}_i \bs{g}_i^t\right\|^2 \\
& \leq \alpha_*^t \frac{N-K}{K} \sum_{i=1}^N \bs{\lambda}_i^2\mathbb{E}\left\|\bs{g}_i^t\right\|^2 + N \sum_{i=1}^N \bs{\lambda}_i^2 \mathbb{E}\left\| \bs{g}_i^t\right\|^2 \\
& = \left(\alpha_*^t \frac{N-K}{K} + N \right)\sum_{i=1}^N \bs{\lambda}_i^2\mathbb{E}\left\|\bs{g}_i^t\right\|^2 \\
& \leq \left(\frac{\alpha_*^t (N-K)}{K} + N\right) W \sum_{i=1}^N \bs{\lambda}_i\mathbb{E}\left\|\bs{g}_i^t\right\|^2 \\
& \leq \left(\frac{\alpha_*^t (N-K)}{K} + N\right) W \eta_l^2 (5R\sigma_l^2 + 30R^2\sum_{i=1}^N \bs{\lambda}_i \mathbb{E}\left\| \nabla f_i(\bs{x}^t) \right\|^2) \\
& \leq \left(\frac{\alpha_*^t (N-K)}{K} + N\right) W \eta_l^2 (5R\sigma_l^2 + 30R^2 (\sigma_g^2 + \mathbb{E}\left\| \nabla f(\bs{x}^t) \right\|^2)). \\
\end{aligned}
$$

Letting $\gamma^t = \left(\frac{\alpha_*^t (N-K)}{K} + N\right) W$, we obtain
$$
\begin{aligned}
(A) & = 5R^2\gamma^t \eta_l^2 (\sigma^2 + \mathbb{E}\left\| \nabla f(\bs{x}^t) \right\|^2).
\end{aligned}
$$

Therefore, we have 
\begin{equation}\label{eq:t2}
\begin{aligned}
T_2 = \mathbb{E}\left[\|\tilde{\boldsymbol{d}}^t\|^2\right] & = \frac{\eta^2}{R^2\eta_l^2} \mathbb{E}\left[\|\boldsymbol{d}^t\|^2\right] \leq \eta_g^2 Q(S^t) + 5\gamma^t \sigma^2 \eta^2 + 5\gamma^t \eta^2 \mathbb{E}\left\| \nabla f(\bs{x}^t) \right\|^2 
\end{aligned}
\end{equation}

\textbf{Putting together.} 
Reorganizing the descent lemma, we obtain
$$
\begin{aligned}
\|\nabla f(\bs{x}^t)\|^2 \leq \frac{2(f(\bs{x}^t) - \mathbb{E}\left[f(\bs{x}^{t+1})\right])}{\eta} + T_1 + L\eta \cdot T_2
\end{aligned}
$$

Then, substituting corresponding terms in \eqref{eq:descent} with \eqref{eq:t1} and \eqref{eq:t2} to finish the descent lemma, we have 
$$
\begin{aligned}
\|\nabla f(\bs{x}^t)\|^2 & \leq \frac{2(f(\bs{x}^t) - \mathbb{E}\left[f(\bs{x}^{t+1})\right])}{\eta} \\
& \quad + \frac{15}{16}\|\nabla f(\bs{x}^t)\|^2 + 10L^2 \sigma^2 \frac{\eta^2}{\eta_g^2} \\
& \quad + L Q(S^t) \eta_g^2 \eta + 5L\gamma^t \eta^3 \sigma^2 + 5L\gamma^t \eta^3 \mathbb{E}\left\| \nabla f(\bs{x}^t) \right\|^2.
\end{aligned}
$$

Since $\eta \leq \frac{1}{8L}$, we have
\begin{equation}
\begin{aligned}
\frac{1}{16}(1 - 10 \gamma^t \eta^2)\|\nabla f(\bs{x}^t)\|^2 & \leq \frac{2(f(\bs{x}^t) - \mathbb{E}\left[f(\bs{x}^{t+1})\right])}{\eta} + \left(L\eta_g^2 Q(S^t) + \frac{5L}{4\eta_g^2}\sigma^2\right)\eta + \frac{5}{8}\gamma^t\sigma^2\eta^2  \\
& \leq \frac{2(f(\bs{x}^t) - \mathbb{E}\left[f(\bs{x}^{t+1})\right])}{\eta} +  \left(Q(S^t) + \frac{5}{4}L\sigma^2\right)\eta + \frac{5}{8}\gamma^t\sigma^2\eta^2,
\end{aligned}
\end{equation}
where we assume $1 \leq \eta_g \leq \max(\sqrt{\frac{1}{L}}, 1)$. Then, we have the final descent lemma as
\begin{equation}\label{eq:descent_org}
\frac{1}{16}(1 - 10 \gamma^t \eta^2)\|\nabla f(\bs{x}^t)\|^2 \leq \frac{2(f(\bs{x}^t) - \mathbb{E}\left[f(\bs{x}^{t+1})\right])}{\eta} +  \left(Q(S^t) + \frac{5}{4}L\sigma^2\right)\eta + \frac{5}{8}\gamma^t\sigma^2\eta^2.
\end{equation}

Then, taking averaging of both sides of \eqref{eq:descent_org} over from time $0$ to $T-1$, we have
$$
\frac{1}{T}\sum_{t=1}^{T} \frac{(1 - 10 \gamma^t \eta^2)}{16} \mathbb{E} \|\nabla f(\bs{x}^t)\|^2 \leq \frac{2(\mathbb{E}\left[f(\bs{x}^0)-f(\bs{x}^{T})\right])}{T\eta} + (\frac{1}{T}\sum_{t=0}^{T-1} Q(S^t) + \frac{5}{4}L\sigma^2)\eta + (\frac{1}{T}\sum_{t=0}^{T-1} \gamma^t) \frac{5}{8}\sigma^2 \eta^2.
$$

Noting that $\eta \leq \frac{1}{8L}$, suppose upper bound $\mathbb{E}[f(\boldsymbol{x}^0) - f(\boldsymbol{x}^{T-1})] \leq M$ and define
$$
\beta_1 = \frac{1}{T}\sum_{t=0}^{T-1} Q(S^t), \quad \beta_2 = \frac{1}{T}\sum_{t=0}^{T-1} \gamma^t, \quad \rho = \min_{t\in [T]} \{\frac{(1 - 10 \gamma^t \eta^2)}{16}\} > 0
$$
we use Lemma~\ref{lemma:constant_stepsize} to tune the stepsize $\eta$ and obtain
$$
\rho \min_{t\in[T]} \mathbb{E} \|\nabla f(\bs{x}^t)\|^2 \leq \left(\frac{2M(4\beta_1 + 5L\sigma^2)}{T}\right)^{\frac{1}{2}}+(5\sigma^2 \beta_2)^{\frac{1}{3}}\left(\frac{2M}{T}\right)^{\frac{2}{3}}+\frac{2LM}{T},
$$
which concludes the proof.
\end{proof}

\section{Detail Proofs of Online Convex Optimization}\label{proof:main}

\subsection{Vanising Hindsight Gap: Proof of Lemma~\ref{bound:hindsight}}\label{proof:bound:hindsight}


\begin{proof}
We first arrange the term (B) in Equation \eqref{eq:regret-dec} as follows,
\begin{equation}\label{def:bound}
\begin{aligned}
    \min_{\bs{p}} \sum_{t=1}^T \ell_t(\bs{p}) -  \sum_{t=1}^T \min_{\bs{p}} \ell_t(\bs{p}) =\min_{\bs{p}}\sum_{t=1}^T \sum_{i=1}^N \frac{\pi_t^2(i)}{\bs{p}_i} -  \sum_{t=1}^T \min_{\bs{p}} \sum_{i=1}^N \frac{\pi_t^2(i)}{\bs{p}_i}.\\
\end{aligned}
\end{equation}

Here, we recall our mild Assumption~\ref{asp:local_convergence},
$$
\pi_*(i) := \lim_{t\rightarrow \infty} \pi_t(i), \; \Pi_* := \sum_{i=1}^N \pi_*(i), \; \forall i \in [N].
$$
Then, denoting $V_T(i) := \sum_{t=1}^T (\pi_t(i) - \pi_*(i))^2$, we bound the cumulative variance over time $T$ per client $i \in [N]$, 
\begin{equation}\label{bound:over-t}
\begin{aligned}
\pi^2_{1:T}(i) = & \sum_{t=1}^T (\pi_*(i) + (\pi_t(i) - \pi_*(i)))^2 \\
\leq & T \cdot \pi^2_*(i) + 2 \pi_*(i) \sum_{t=1}^T |\pi_t(i) - \pi_*(i)| + \sum_{t=1}^T (\pi_t(i) - \pi_*(i))^2 \\
\leq & T \cdot \pi^2_*(i) + 2 \pi_*(i)\sqrt{T\cdot V_T(i)} + V_T(i) \\
= & T \left( \pi_*(i) + \sqrt{\frac{V_T(i)}{T}} \right)^2.
\end{aligned}
\end{equation}
Using the Lemma \ref{lemma:optimal} and non-negativity of feedback we have, 
\begin{equation}\label{obj:minp}
    \min_{\bs{p}} \sum_{i=1}^N \frac{\pi^2_t(i)}{\bs{p}_i} = \frac{(\sum_{i=1}^N \pi_t(i))^2}{K}.
\end{equation}

We obtain the upper bound of the first term in Equation \eqref{def:bound},
\begin{equation}\label{bound:term1}
\begin{aligned}
\min_{\bs{p}} \sum_{t=1}^T \sum_{i=1}^N \frac{\pi^2_t(i)}{\bs{p}_i} = & \min_{\bs{p}} \sum_{i=1}^N \frac{\pi^2_{1:T}(i)}{\bs{p}_i}
= \frac{\Big(\sum_{i=1}^N \sqrt{\pi^2_{1:T}(i)}\Big)^2}{K} \\
\leq & \frac{T}{K} \Bigg(\sum_{i=1}^N \pi_*(i) + \sum_{i=1}^N \sqrt{\frac{V_T(i)}{T}} \Bigg)^2 \\
= &  \frac{T}{K} \left(\Pi_*^2 + 2 \Pi_* \sum_{i=1}^N \sqrt{\frac{V_T(i)}{T}} + \Big(\sum_{i=1}^N \sqrt{\frac{V_t(i)}{T}}\Big)^2\right),
\end{aligned}
\end{equation}
where we use Lemma \ref{lemma:optimal} in the second line, and Equation \eqref{bound:over-t} in the third line.

Then, we bound the second term in Equation \eqref{def:bound}:
\begin{equation}\label{bound:term2}
\begin{aligned}
\Pi_*^2 = & \sum_{i=1}^N \pi^2_*(i) 
\leq \left(\frac{1}{T} \sum_{t=1}^T \sum_{i=1}^N \pi_t(i)\right)^2 \leq \frac{1}{T} \sum_{t=1}^T (\sum_{i=1}^N \pi_t(i))^2 \\
=& \frac{K}{T} \sum_{t=1}^T \min_{\bs{p}} \sum_{i=1}^N \frac{\pi^2_t(i)}{\bs{p}_i}, 
\end{aligned}
\end{equation}
where the first inequality uses the average assumption, the third inequality uses Jensen's inequality, and the last inequality uses Equation \eqref{obj:minp}.

Overall, we combine the results in Equation \eqref{bound:term1} and \eqref{bound:term2}, and conclude the proof:
\begin{equation}
\min_{\bs{p}}\sum_{t=1}^T \ell_t(\bs{p}) -  \sum_{t=1}^T \min_{\bs{p}} \ell_t(\bs{p}) \leq \frac{T}{K} \left(\sum_{i=1}^N \sqrt{\frac{V_T(i)}{T}} \right)\left(2 \Pi_* + \sum_{i=1}^N \sqrt{\frac{V_t(i)}{T}}\right).
\end{equation}
\end{proof}

\subsection{Regret of Full Information}\label{apd:full_info}

\begin{theorem}[Static regret with full information]\label{theorem:full_ftrl} Under Assumptions~\ref{asp:local_convergence}, sampling a batch of clients with an expected size of $K$, and setting $\gamma = G^2$, the FTRL scheme in \eqref{obj:ol_ftrl} yields the following regret,
\begin{equation}\label{eq:full_feedback}
\begin{aligned}
    \sum_{t=1}^T \ell_t(\bs{p}^t) - \min_{\bs{p}}\sum_{t=1}^T \ell_t(\bs{p}) \leq  \left(\frac{22NG}{\bar{z}} + \frac{2\sqrt{6} NG}{K}\right)\sum_{i=1}^N \sqrt{\pi^2_{1:T}(i)} + \frac{22NG^2}{\bar{z}},
\end{aligned}
\end{equation}
where we note the cumulative feedback $\sqrt{\pi^2_{1:T}(i)} \leq \mathcal{O}(\sqrt{T})$ following Assumption~\ref{asp:local_convergence}.
\end{theorem}


\begin{proof}
We considering a restricted probability space $\Delta = \{\bs{p} \in \mathbb{R}^N | \bs{p}_i \geq p_{\text{min}}, \sum_{i=1}^N \bs{p}_i = K, \forall i \in [N]\}$ where $p_{\text{min}} \leq K/N$. Then, we decompose the regret,
\begin{equation}
    \text{Regret}_{\text{FTRL}}(T) = \underbrace{\sum_{t=1}^T \ell_t(\bs{p}^t) - \min_{p\in\Delta}\sum_{t=1}^T \ell_t(\bs{p})}_{(A)} + \underbrace{\min_{p\in\Delta}\sum_{t=1}^T \ell_t(\bs{p}) - \min_{\bs{p}}\sum_{t=1}^T \ell_t(\bs{p})}_{(B)}.
\end{equation}
We separately bound the above terms in this section. The bound of (A) is related to the stability of the online decision sequence by playing FTRL, which is given in Lemma \ref{bound:ftrl-stab}. Term (B) is bounded by the minimal results of directing calculation. 

\textbf{Bounding (A)}. Without loss of generality, we introduce the stability of the online decision sequence from FTRL to variance function $\ell$ as shown in the following lemma\citep{kalai2005efficient} (similar proof can also be found in \citep{hazan201210, shalev2012online}).
\begin{lemma}\label{bound:ftrl-stab}
Let $\mathcal{K}$ be a convex set and $\mathcal{R}:\mathcal{K}\mapsto\mathbb{R}$ be a regularizer. Given a sequence of functions $\{\ell_t\}_{t\in[T]}$ defined over $\mathcal{K}$, then setting $\bs{p}^t = \arg \min_{\bs{p}\in \mathbb{R}^N} \sum_{\tau=1}^{t-1} 
\ell_\tau(\bs{p}) + \mathcal{R}(\bs{p})$ ensures,
$$
\sum_{t=1}^T \ell_t(\bs{p}^t) - \sum_{t=1}^T \ell_t(\bs{p}) \leq \sum_{t=1}^T (\ell_t(\bs{p}^t) - \ell_t(\bs{p}^{t+1})) + (\mathcal{R}(\bs{p}) - \mathcal{R}(\bs{p}^1)), \forall \bs{p} \in \mathcal{K}.
$$
\end{lemma}

We note that $\mathcal{R}(\bs{p}) =  \sum_{i=1}^N \gamma/\bs{p}_i$ in our work. Furthermore, $\mathcal{R}(\bs{p})$ is non-negative and bounded by $N \gamma / p_{\text{min}}$ with $p\in \Delta$. Thus, the above lemma incurs,
\begin{equation}\label{bound:ftrl-pre}
\sum_{t=1}^T \ell_t(\bs{p}^t) - \sum_{t=1}^T \ell_t(\bs{p}) \leq \underbrace{\sum_{t=1}^T (\ell_t(\bs{p}^t) - \ell_t(\bs{p}^{t+1})) }_{\text{Bounded Below}}+ \frac{N\gamma}{p_{\text{min}}}.
\end{equation}

To simply the following proof, we assume that $0 < \pi_1(t) \leq  \pi_2(t) \leq \dots \leq \pi_N(t), t \in [T]$ to satisfies Lemma~\ref{obj:restrict} without the loss of generality. The stability relies on the evolution of cumulative feedback $\pi_{1:t}^2(i)$ and hence relies on the index in solution $l_1, l_2$ according to Lemma~\ref{lemma:optimal}. Following the Lemma~\ref{obj:restrict}, we have 
\begin{equation}
    \bs{p}_i^t = \begin{cases}
        1, & \text{if}\; i\geq l_2^t ,\\
        z_t \frac{\sqrt{\pi_{1:t-1}^2(i) + \gamma}}{c_t}, & \text{if}\; i \in (l_1^t, l_2^t), \\
        p_{\text{min}}, & \text{if}\; i\leq l_1^t, \\
    \end{cases}
\end{equation}
where $z_t = K - N + l_2^t - l_1^t \cdot p_{\text{min}} \leq K $ and $c_t = \sum_{i\in(l_1^t, l_2^t)} \sqrt{\pi_{1:t}^2(i) +\gamma} \leq \sum_{i=1}^N \sqrt{\pi_{1:t}^2(i) +\gamma}$ is the normalization factor . Then, we investigate the first term in the above inequality,
$$
\begin{aligned}
\sum_{t=1}^T (\ell_t(\bs{p}^t) - \ell_t(\bs{p}^{t+1})) & \leq \sum_{t=1}^T\sum_{i=1}^N \pi^2_t(i) \cdot \Bigg(\frac{1}{\bs{p}_i^t} - \frac{1}{\bs{p}_i^{t+1}}\Bigg).\\
\end{aligned}
$$

\textbf{Remark}. According to the above inequality, we note that the stability of online convex optimization is highly related to the changing probability. We can have a trivial upper bound $\sum_{t=1}^T (\ell_t(\bs{p}^t) - \ell_t(\bs{p}^{t+1})) \leq \sum_{t=1}^T\sum_{i=1}^N \pi^2_t(i) \cdot (1/p_{\text{min}}-1)$, which indicates that the stability is restricted by $p_{\text{min}}$. Solving the sampling probability requires sorting cumulative feedbacks $\pi_{1:t}^2(i)$, the combinations of client-index and $\bs{p}_i^t$ are dynamic. Hence, directly bounding the above equation generally can be difficult. To obtain a tighter bound for FTRL, we investigate the possible 

\begin{lemma}\label{lemma:stability}
Assuming that $\bs{p}_i^t < \bs{p}_i^{t+1}$, for all $i\in[N], t\in[T-1]$, the upper bound of $\Bigg(\frac{1}{\bs{p}_i^t} - \frac{1}{\bs{p}_i^{t+1}}\Bigg)$ is given by:
\begin{equation}\label{eq:stability}
0 \leq \Bigg(\frac{1}{\bs{p}_i^t} - \frac{1}{\bs{p}_i^{t+1}}\Bigg) \leq \frac{1}{\min(z_{t},z_{t+1})}\Bigg(\frac{c_t}{\sqrt{\pi_{1:t-1}^2(i) + \gamma}} - \frac{c_{t+1}}{\sqrt{\pi_{1:t}^2(i) + \gamma}}\Bigg).
\end{equation}
\end{lemma}
\begin{proof}
For all $t\in[T]$, we have cumulative feedbacks $\pi_{1:t-1}(i), \; i\in[N]$ on the server. The server is able to compute results \eqref{eq:time_results_p}. As we are interested in the upper bound, we assume $\bs{p}_i^t < \bs{p}_i^{t+1}$ and discuss the cases below:
\begin{itemize}
\item \textbf{Case 1}: letting $(\bs{p}_i^t, \bs{p}_i^{t+1}) = (p_{\text{min}}, z_{t+1} \frac{\sqrt{\pi_{1:t}^2(i) + \gamma}}{c_{t+1}})$, we have
$$
\begin{aligned}
\frac{1}{\bs{p}_i^t} - \frac{1}{\bs{p}_i^{t+1}} & = \frac{1}{p_{\text{min}}} - \frac{c_{t+1}}{z_{t+1}\sqrt{\pi_{1:t}^2(i) + \gamma}} \\
& \leq \frac{c_t}{z_{t}\sqrt{\pi_{1:t-1}^2(i) + \gamma}} - \frac{c_{t+1}}{z_{t+1}\sqrt{\pi_{1:t}^2(i) + \gamma}},\\
& \leq \frac{1}{\min(z_{t},z_{t+1})}\left(\frac{c_t}{\sqrt{\pi_{1:t-1}^2(i) + \gamma}} - \frac{c_{t+1}}{\sqrt{\pi_{1:t}^2(i) + \gamma}}\right),
\end{aligned}
$$
where the second inequality uses \eqref{obj:optimalp} indicating $p_{\text{min}} \geq z_t \frac{\sqrt{\pi_{1:t-1}^2(i) + \gamma}}{c_t}$.

\item \textbf{Case 2}: letting $(\bs{p}_i^t, \bs{p}_i^{t+1}) = (z_{t} \frac{\sqrt{\pi_{1:t-1}^2(i) + \gamma}}{c_{t}}, z_{t+1} \frac{\sqrt{\pi_{1:t}^2(i) + \gamma}}{c_{t+1}})$, \eqref{eq:stability} naturally holds.

\item \textbf{Case 3}: letting $(\bs{p}_i^t, \bs{p}_i^{t+1}) = (z_{t} \frac{\sqrt{\pi_{1:t-1}^2(i) + \gamma}}{c_{t}}, 1)$, we can know that $1 \leq z_{t+1} \frac{\sqrt{\pi_{1:t}^2(i) + \gamma}}{c_{t+1}}$ by \eqref{obj:optimalp} and prove the conclusion analogous to case 1.

\item \textbf{Case 4}: analogous to the case 1 and 3, letting $(\bs{p}_i^t, \bs{p}_i^{t+1}) = (p_{\text{min}}, 1)$, \eqref{eq:stability} naturally holds.
\end{itemize}
Summarizing all cases to conclude the proof.
\end{proof}

Using Lemma~\ref{lemma:stability}, we are ready to bound the stability of the online decision sequence:
$$
\begin{aligned}
\sum_{t=1}^T (\ell_t(\bs{p}^t) - \ell_t(\bs{p}^{t+1})) & = \sum_{t=1}^T\sum_{i=1}^N \pi^2_t(i) \cdot \Bigg(\frac{c_t}{z_t \sqrt{\pi_{1:t-1}^2(i) + \gamma}} - \frac{c_{t+1}}{z_{t+1}\sqrt{\pi_{1:t}^2(i) + \gamma}}\Bigg)\\
& \leq \sum_{t=1}^T\sum_{i=1}^N \frac{\pi^2_t(i) \cdot c_t}{\min(z_t, z_{t+1})} \cdot \Bigg(\frac{1}{\sqrt{\pi_{1:t-1}^2(i) + \gamma}} - \frac{1}{\sqrt{\pi_{1:t}^2(i) + \gamma}}\Bigg) & \text{$\triangleright c_t \leq c_{t+1}$}\\
& \leq \sum_{t=1}^T\sum_{i=1}^N \frac{\pi^2_t(i) \cdot \tilde{c_t}}{\min(z_t, z_{t+1})\sqrt{\pi^2_{1:t}(i) + \gamma}} \cdot \Bigg(\sqrt{1 + \frac{\pi^2_t(i)}{\pi^2_{1:t-1} (i) + \gamma}} - 1\Bigg)\\
& \leq \frac{\tilde{c_T}}{2} \sum_{t=1}^T\sum_{i=1}^N \frac{1}{\min(z_t, z_{t+1})}\frac{\pi_t(i)^4}{\sqrt{\pi^2_{1:t}(i) + \gamma}\cdot (\pi^2_{1:t-1} (i) + \gamma)}, & \text{$\triangleright \sqrt{1+x}-1 \leq \frac{x}{2}$}\\
\end{aligned}
$$
where the third line uses definition $c_t \leq \tilde{c_t} = \sum_{i=1}^N \sqrt{\pi_{1:t}^2(i) +\gamma}$.

Letting $\gamma = G^2$, we have that $\pi^2_{1:t}(i) \leq \pi^2_{1:t-1} (i) + \gamma$ and $\sqrt{\pi^2_{1:t}(i)} \leq \sqrt{\pi^2_{1:t}(i) + \gamma}$. We define $\bar{z} = \min \{z_t\}_{t=1}^{T}$ and conclude the bound,
\begin{equation}\label{bound:ftrl-al}
\begin{aligned}
\sum_{t=1}^T (\ell_t(\bs{p}^t) - \ell_t(\bs{p}^{t+1})) \leq & \frac{\tilde{c_T}}{2} \sum_{t=1}^T\sum_{i=1}^N \frac{\pi_t(i)^4}{(\pi^2_{1:t}(i))^\frac{3}{2}} \\
= & G \cdot \frac{\tilde{c_T}}{2 \bar{z}} \sum_{i=1}^N \sum_{t=1}^T\frac{(\pi_t(i)/G)^4}{((\pi_{1:t}(1)/G)^2)^\frac{3}{2}} & \text{$\triangleright \text{Lemma~\ref{lemma:sum_a}}$} \\
\leq & \frac{22 N G}{\bar{z}} \sum_{i=1}^N \sqrt{\pi_{1:T}^2(i) +G^2} & \text{$\triangleright \text{Definition of } \tilde{c_T}$}\\
\leq & \frac{22 N G}{\bar{z}} \sum_{i=1}^N \left(\sqrt{\pi_{1:T}^2(i)} + G\right)
\end{aligned}
\end{equation}

Finally, we can get the final bound of (A) by plugging~\eqref{bound:ftrl-al} into~\eqref{bound:ftrl-pre} and summarizing as follows,
$$
\sum_{t=1}^T \ell_t(\bs{p}^t) - \sum_{t=1}^T \ell_t(\bs{p}) \leq \frac{22 N G}{\bar{z}} \sum_{i=1}^N \left(\sqrt{\pi_{1:T}^2(i)} + G\right) + \frac{NG^2}{p_{\text{min}}}.
$$

\textbf{Bounding (B)}. Using Corollaries~\ref{cor:min} and \ref{cor:pmin-bound}, we bound the term (B) as follows,
\begin{equation}\label{bound:ftrl-b}
\begin{aligned}
& \min_{p\in\Delta}\sum_{t=1}^T \ell_t(\bs{p}) - \min_{\bs{p}}\sum_{t=1}^T \ell_t(\bs{p}) \\
\leq & \frac{K(\sum_{i={1}}^{N} \sqrt{\pi^2_{1:T}(i)} )^2}{(K-Np_{\text{min}})^2} - \frac{(\sum_{i=1}^N \sqrt{\pi^2_{1:T}(i)})^2}{K} \\
\leq & \Big(\frac{K}{(K-Np_{\text{min}})^2} - \frac{1}{K}\Big) \cdot \Bigg(\sum_{i=1}^N \sqrt{\pi^2_{1:T}(i)}\Bigg)^2 \\
\leq & \frac{6Np_{\text{min}}}{K^2} \cdot \Bigg(\sum_{i=1}^N \sqrt{\pi^2_{1:T}(i)}\Bigg)^2
\end{aligned}
\end{equation}
In the last line, we use the fact that $\frac{1}{(1-x)^2} - 1 \leq 6x$ for $x\in [0, 1/2]$. Hence, we scale the coefficient 
$$
\frac{K}{(K-Np_{\text{min}})^2} - \frac{1}{K} = \frac{1}{K}\Big[ \frac{1}{(1- Np_{\text{min}}/K)^2}-1\Big] \leq \frac{6Np_{\text{min}}}{K^2},
$$
where we let $p_{\text{min}} \leq K/(2N)$. 

\textbf{Summary}. Setting $\gamma = G^2$, and combining the bound in~\eqref{bound:ftrl-pre} and~\eqref{bound:ftrl-b}, we have,
\begin{equation}
\begin{aligned}
    & \text{Regret}_{\text{FTRL}}(T) = \sum_{t=1}^T \ell_t(\bs{p}^t) - \min_{\bs{p}}\sum_{t=1}^T \ell_t(\bs{p}) \\
    \leq & \frac{22 N G}{\bar{z}} \sum_{i=1}^N \left(\sqrt{\pi_{1:T}^2(i)} + G\right) + \frac{NG^2}{p_{\text{min}}} + \frac{6Np_{\text{min}}}{K^2} \cdot \Bigg(\sum_{i=1}^N \sqrt{\pi^2_{1:T}(i)}\Bigg)^2.
\end{aligned}
\end{equation}

The $p_{\text{min}}$ is only relevant for the theoretical analysis. Hence, the choice of it is arbitrary, and we can set it to $p_{\text{min}} = \min\Big\{K/(2N), G K /(\sqrt{6}\sum_{i=1}^N\sqrt{\pi^2_{1:T}(i)})  \Big\}$ which turns the upper bound to the minimal value. Hence, we yield the final bound of FTRL in the end,
\begin{equation}
 \sum_{t=1}^T \ell_t(\bs{p}^t) - \min_{\bs{p}}\sum_{t=1}^T \ell_t(\bs{p}) \leq \left(\frac{22NG}{\bar{z}} + \frac{2\sqrt{6} NG}{K}\right)\sum_{i=1}^N \sqrt{\pi^2_{1:T}(i)} + \frac{22NG^2}{\bar{z}}.
\end{equation}
\end{proof}

\subsection{Expected Regret of Partial Feedback: Proof of Theorem~\ref{bound:soft-regret}}\label{proof:bandit}

\begin{proof}
Using the property of unbiasedness, we have
\begin{equation}
\begin{aligned}
& \min_{\bs{p}} \mathbb{E}[\sum_{t=1}^T \ell_t(\tilde{\bs{p}}^t) - \sum_{t=1}^T \ell_t(\bs{p})] \\
= & \min_{\bs{p}} \mathbb{E}[\sum_{t=1}^T \tilde{\ell}_t(\tilde{\bs{p}}^t) - \sum_{t=1}^T \tilde{\ell}_t(\bs{p})] \\
= & \underbrace{\mathbb{E}\Big[ \sum_{t=1}^T \tilde{\ell}_t(\tilde{\bs{p}}^t) - \sum_{t=1}^T \tilde{\ell}_t(\bs{p}^t) \Big]}_{(A)} + \underbrace{\min_{\bs{p}} \mathbb{E}\Big[ \sum_{t=1}^T \tilde{\ell}_t(\bs{p}^t) - \sum_{t=1}^T \ell_t(\bs{p}) \Big]}_{(B)}.
\end{aligned}
\end{equation}

\textbf{Bounding (A)}. We recall that $\tilde{\bs{p}}_i^t \geq \frac{\theta K}{N}$ for all $t\in[T], i \in[N]$ due to the mixing. Therefore, $\bs{p}_i^t \geq K/N$ implies $\tilde{\bs{p}}_i^t \geq K/N$. Thus, we have
$$
 \frac{1}{\tilde{\bs{p}}_i^t} - \frac{1}{\bs{p}_i^t} = \theta \cdot \frac{\bs{p}_i^t - \frac{K}{N}}{\tilde{\bs{p}}_i^t\bs{p}_i^t} \leq \theta \cdot \frac{\bs{p}_i^t}{\tilde{\bs{p}}_i^t\bs{p}_i^t} = \frac{\theta}{\tilde{\bs{p}}_i^t} \leq \theta\cdot\frac{N}{K}.
$$
Moreover, if $\bs{p}_i^t \leq K/N$, the above inequality still holds. We extend the (A) as follows,
$$
\begin{aligned}
(A) & := \mathbb{E}\Big[ \sum_{t=1}^T \tilde{\ell}_t(\tilde{\bs{p}}^t) - \sum_{t=1}^T \tilde{\ell}_t(\bs{p}^t) \Big] \\
& = \mathbb{E}\Big[\sum_{t=1}^T \sum_{i=1}^N \tilde{\pi}^2_t(i) \Big( \frac{1}{\tilde{\bs{p}}_i^t} - \frac{1}{\bs{p}_i^t}\Big)\Big] \\
& \leq \theta \cdot \frac{N}{K} \cdot \mathbb{E}\Big[ \sum_{t=1}^T \sum_{i=1}^N \tilde{\pi}^2_t(i) \Big] \\
& \leq \frac{\theta G^2N^2}{K}T,
\end{aligned}
$$
where we use $\mathbb{E}[\tilde{\pi}^2_t(i)] = \pi^2_t(i) \leq G^2$.

\textbf{Bounding (B)}. We note that $\bs{p}^t$ is the decision sequence playing FTRL with the mixed cost functions. Thus, we combine the mixing bound of feedback (i.e., $\tilde{\pi}^2_t(i) \leq \frac{G^2 N}{\theta K}$) and Theorem~\ref{theorem:full_ftrl}. Replacing $G^2$ with $G^2 \frac{N}{\theta K}$, we get

\begin{equation}\label{eq:best_gamma}
\sum_{t=1}^T \tilde{\ell}_t(\bs{p}^t) - \min_{\bs{p}} \sum_{t=1}^T \tilde{\ell}_t(\bs{p}) \leq \left(\frac{22N^{\frac{3}{2}}G}{\bar{z}\sqrt{\theta K}} + \frac{2\sqrt{6} N^{\frac{3}{2}}G}{\sqrt{\theta K^3}}\right)\mathbb{E}\left[\sum_{i=1}^N \sqrt{\tilde{\pi}^2_{1:T}(i)}\right] + \frac{22G^2N^2}{\bar{z} \theta K}.
\end{equation}

\textbf{Summary}. Using Jensen's inequality, we have $\mathbb{E}\big[\sum_{i=1}^N \sqrt{\tilde{\pi}^2_{1:T}(i)}\big] \leq \sum_{i=1}^N \sqrt{\mathbb{E}[\tilde{\pi}^2_{1:T}(i)]} = \sum_{i=1}^N \sqrt{\pi^2_{1:T}(i)}$. Finally, we can get the upper bound of the regret in partial-bandit feedback,
\begin{equation}\label{eq:best_theta}
\begin{aligned}
N^2 \cdot \min_{\bs{p}} \mathbb{E}[\sum_{t=1}^T \ell_t(\tilde{\bs{p}}^t) - \sum_{t=1}^T \ell_t(\bs{p})] & \leq \frac{\theta G^2}{K}T + \left(\frac{22G}{\bar{z}\sqrt{\theta N K}} + \frac{2\sqrt{6} G} {\sqrt{\theta N K^3}}\right)\mathbb{E}\left[\sum_{i=1}^N \sqrt{\tilde{\pi}^2_{1:T}(i)}\right] + \frac{22G^2}{\bar{z} \theta K} \\
&\leq \frac{\theta G^2 }{K} T + \left(\frac{22N^{\frac{1}{2}}G^2}{\bar{z}\sqrt{\theta K}} + \frac{2\sqrt{6} N^{\frac{1}{2}}G^2} {\sqrt{\theta K^3}}\right)\sqrt{T} + \frac{22G^2}{\bar{z} \theta K},
\end{aligned}
\end{equation}
where the last line uses the bound $\sum_{i=1}^N \sqrt{\pi^2_{1:T}(i)} \leq NG\sqrt{T}$. Now, we can optimize the upper bound of regret in terms of $\theta$. Notably, $\theta$ is independent on $T$ and we set $\theta = (\frac{N}{TK})^{\frac{1}{3}}$ to get the minimized bound. Additionally, we are pursuing an expected regret, which is $\text{Regret}_{(S)}(T)$ in the original definition in~\eqref{eq:regret-dec}. Using the unbiasedness of the mixed estimation and modified costs, we can obtain the final bound:
$$
\begin{aligned}
N^2 \cdot \mathbb{E}[\text{Regret}_{(S)}(T)] & = \mathbb{E}[\sum_{t=1}^T \ell_t(\tilde{\bs{p}}^t) - \min_{\bs{p}}\sum_{t=1}^T \ell_t(\bs{p})] \\ 
& =  \mathbb{E}[\sum_{t=1}^T \ell_t(\tilde{\bs{p}}^t) - \min_{\bs{p}}\sum_{t=1}^T \tilde{\ell}_t(\bs{p})] +  \mathbb{E}[\min_{\bs{p}}\sum_{t=1}^T \tilde{\ell}_t(\bs{p}) - \min_{\bs{p}}\sum_{t=1}^T \ell_t(\bs{p})] \\
& \leq \mathcal{O}\big(N^{\frac{1}{3}}T^{\frac{2}{3}}/K^{\frac{4}{3}}\big) + \mathbb{E}[\min_{\bs{p}}\sum_{t=1}^T \tilde{\ell}_t(\bs{p}) - \min_{\bs{p}}\sum_{t=1}^T \ell_t(\bs{p})]  \\
& \leq \tilde{\mathcal{O}}\big(N^{\frac{1}{3}}T^{\frac{2}{3}}/K^{\frac{4}{3}}\big),
\end{aligned}
$$
where the last inequality uses Lemma~\ref{bound:hindsight}, and the conclusion in Theorem 8~\citep{borsos2018online}. It proves the second term induces an additional log term to the final bound.

\textbf{Remark.} Baseline works have additional averaging coefficient $\frac{1}{N^2}$ in their final bound. This is because they consider the weights $\bs{\lambda}=1/N$ in stochastic optimization, while we include the $\lambda$ for clients' weights in federated optimization. To align with them, we omit the coefficient of $N^2$ and report the final bound for $\mathbb{E}[\text{Regret}_{(S)}(T)]$, as $N^2$ can be absorbed by excluding the $\bs{\lambda}$ from client feedback function $\pi(\cdot)$.
\end{proof}

\section{Further Discussions}\label{sed:discussion}

\subsection{A Sketch of Proof with Client Stragglers}\label{dis:fail}

We note the possibility that some clients are unavailable to participants due to local failure or being busy in each round. To extend our analysis to the case, we assume there is a known distribution of client availability $\mathcal{A}$ such that a subset $\mathcal{A}^t \sim \mathcal{A}$ of clients are available at the $t$-th communication round. Let $\bs{q}_i = \text{Prob}(i\in \mathcal{A}^t)$ denote the probability that client $i$ is available at round $t$. Based on the setting, we update the definition of estimation $\bs{d}^t$:
$$
\bs{d}^t := \sum_{i\in S^t \subseteq \mathcal{A}^t} \frac{\bs{\lambda}_i \bs{g}_i^t}{\bs{q}_i \bs{p}^t_i},
$$
where $S^t \subseteq \mathcal{A}^t$ indicates that we can only sample from available set. Then, we apply the estimation to variance and obtain the following target:
$$
\text{Regret}(T) = \sum_{t=1}^T \sum_{i=1}^N \frac{\pi_t^2(i)}{\bs{q}_i\bs{p}_i} - \sum_{t=1}^T \min_{\bs{p}} \sum_{i=1}^N \frac{\pi_t^2(i)}{\bs{q}_i\bs{p}_i}.
$$
Analogous to our analysis in Appendix~\ref{proof:main}, we could obtain a similar bound of the above regret that considers the availability.

\subsection{Differences between biased client sampling methods}~\label{sec:diff}
This section discusses the main differences between unbiased client sampling and biased client sampling methods. The proposed K-Vib sampler is an unbiased sampler for the first-order gradient of objective~\ref{eq:obj}. Recent biased client sampling methods include Power-of-Choice (POC)~\citep{cho2020client} and DivFL~\citep{balakrishnan2022diverse}. Concretely, POC requires all clients to upload local empirical loss as prior knowledge and selects clients with the largest empirical loss. DivFL builds a submodular based on the latest gradient from clients and selects clients to approximate all client information. Therefore, these client sampling strategies build a biased gradient estimation that may deviate from a fixed global goal.

FL with biased client sampling methods, such as POC and DivFL, can be considered dynamic re-weighting algorithms adjusting $\bs{p}_i$. Analogous to the \eqref{eq:obj}, the basic objective of FL with biased client sampling methods can be defined as follows~\citep{li2020federated, balakrishnan2022diverse, cho2020client}:
\begin{equation}
    \min_{\bs{x} \in \mathcal{X}} f(\bs{x}) := \sum_{i=1}^N\bs{p}_i f_i(\bs{x}) := \sum_{i=1}^N\bs{p}_i \mathbb{E}_{\xi_i \sim \mathcal{D}_i}[F_i(\bs{x}, \xi_i)],
\end{equation}
where $\bs{p}$ is the probability simplex, and $\bs{p}_i$ is the probability of client $i$ being sampled. The gradient estimation is defined as $\bs{g}^t = \frac{1}{K} \sum_{i \in S^t} \bs{g}_i$ accordingly. The targets of biased FL client sampling are determined by the sampling probability $\bs{p}$ as a replacement of $\bs{\lambda}$ in the original FedAvg objective~\ref{eq:obj}. Typically, the value of $\bs{p}$ is usually dynamic and implicit.

\subsection{Theoretical Comparison with OSMD}\label{sec:comparision_osmd}

The K-Vib sampler proposed in this paper is orthogonal with the recent work OSMD sampler ~\cite{zhao2022adaptive}\footnote{we refer to the latest version \url{https://arxiv.org/pdf/2112.14332.pdf}} in theoretical contribution. We justify our points below:

a) According to Equations (6) and (7) in OSMD, it proposes an online mirror descent procedure that optimizes the additional estimates to replace the mixing strategy in Vrb~\cite{borsos2018online}. The approach can be also utilized as an alternative method in \eqref{eq:utilize}.

b) The improvement of the K-Vib sampler is obtained from the modification of the sampling procedure. In contrast, the OSMD still follows the conventional RSP, as we discussed in Lemma~\ref{lema:variance}. Hence, our theoretical findings of applying the ISP in adaptive client sampling can be transferred to OSMD as well.

In short, the theoretical improvement of our work is independent of the OSMD sampler. And, our insights about utilizing the ISP can be used to improve the OSMD sampler. Meanwhile, the OSMD also suggests future work for the K-Vib sampler in optimizing the additional estimates procedure instead of mixing.

\section{Experiments Details}\label{app:experiments}

\begin{figure}[t]
\centering
\includegraphics[width=0.85\linewidth]{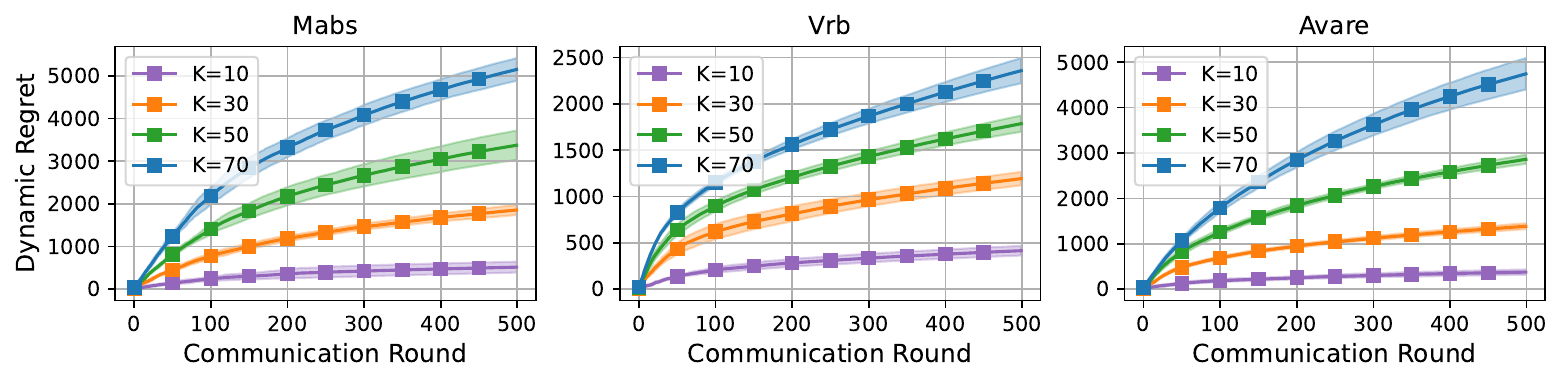}
\caption{Regret of baseline algorithms with 
 different $K$}\label{fig:baseline_effect_k}
\end{figure}

\begin{figure}[t]
\centering
\includegraphics[width=0.85\linewidth]{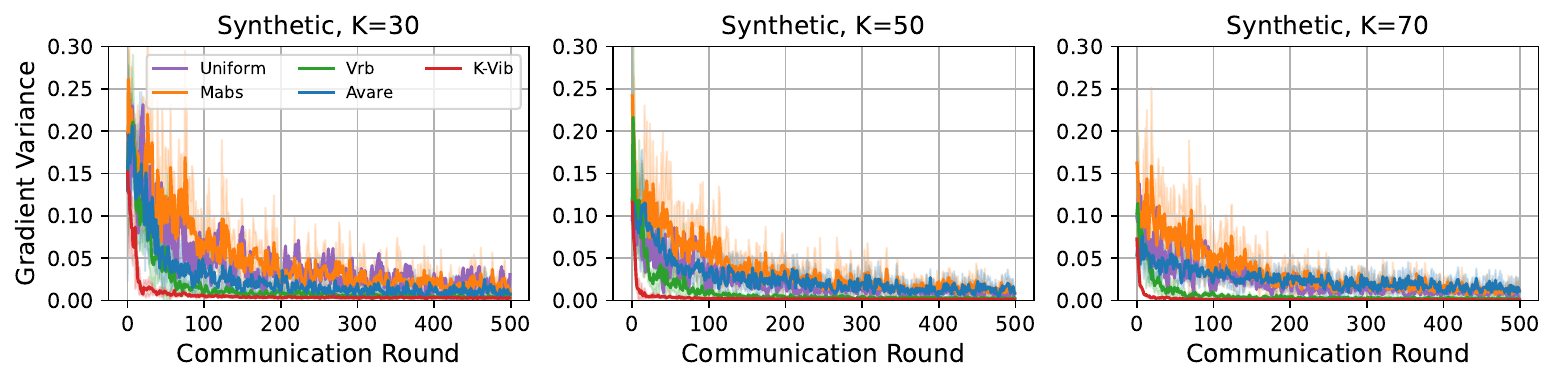}
\caption{Gradient variance with 
 different $K$}\label{fig:variance_reduction_K}
\end{figure}

The experiment implementations are supported by \textit{FedLab} framework~\citep{smile2021fedlab}.  We provide the missing experimental details below:

\textbf{Hyperparameters Setting}. All samplers have an implicit value $G$ related to the hyperparameters. We set $G=0.01$ for the Synthetic dataset task and $G=0.1$ for FEMNIST tasks. We set $\eta=0.4$ (stability hyperparameter) for Mabs~\citep{salehi2017stochastic} as suggested by the original paper. Vrb~\cite{borsos2018online} also utilize mixing strategy $\theta = (N/T)^\frac{1}{3}$ and regularization $\gamma=G^2*N/\theta$. For the case that $N>T$ in FEMNIST tasks, we set $\theta=0.3$ following the official source code\footnote{https://github.com/zalanborsos/online-variance-reduction}. For Avare~\cite{el2020adaptive}, we set $p_{\text{min}}= \frac{1}{5N}$, $C=\frac{1}{\frac{1}{N}-p_{\text{min}}}$ and $\delta=1$ for constant-stepsize as suggested in Appendix D of original paper. For the K-Vib sampler, we set $\theta = (\frac{N}{TK})^{1/3}$ and $\gamma=G^2\frac{N}{K\theta}$. We also fix $\gamma$ and $\theta=0.3$ for our sensitivity study in Figure~\ref{fig:sensitivity}.

\textbf{Baselines with budget $K$.} Our theoretical results in Theorem~\ref{bound:soft-regret} and empirical results in Figure~\ref{fig:effct_k} reveal a key improvement of our work, that is, the linear speed up in online convex optimization. In contrast, we provide additional experiments with the different budget $K$ in Figure~\ref{fig:baseline_effect_k}. Baseline methods do not preserve the improvement property respecting large budget $K$ in adaptive client sampling for variance reduction. Moreover, with the increasing communication budget $K$, the optimal sampling value is decreasing. As a result, the regret of baselines increases in Figure~\ref{fig:baseline_effect_k}, indicating the discrepancy to the optimal is enlarged.

\section{Efficient Implementation}\label{sec:efficient}

In experiments, we do not find a heavy computation time increase compared to baselines as our experiments only involve thousands of clients. To guarantee practical usage for large-scale systems, we present efficient implementation details of K-Vib. 

We can maintain a sorted list of cumulative local weights [$\omega(1), \omega(2), \dots, \omega(N)$] such that $\omega(i) \leq \omega(j), \;\forall i,j \in [N]$ in Algorithm~\ref{alg:sampler}. For each communication round, the server receives feedback values as a list $[\pi_t (j)], \forall j \in S^t$. Then, the server will traverse the feedback list. For each element in the list, the server conducts two main steps as below:
\begin{itemize}
    \item Step 1: For each $j \in S^t$, server computes estimates $\tilde{\omega}(j) = \omega(j) + \pi^2_t(j)/ p_j^t$. Then, the server uses \emph{binary-search} to find the index $k$ such that $\omega(k) \leq \tilde{\omega}(j) < \omega(k+1)$ in the cumulative local weights.
    \item Step 2: Then, server update $\omega(j) = \tilde{\omega}(j)$ and move the position of $\omega(j)$ behind $\omega(k)$ to update the weights sequence.
\end{itemize}

This implementation implements a time complexity of $\mathcal{O}(T \cdot K \cdot \log N)$, where $T$ is the communication round, $K$ is the communication budget, and $N$ is the number of clients. For each communication round $t \in [T]$, the server updates $K$ times of the list with each time cost $\mathcal{O}(\log N)$ to conduct one binary search.

\end{document}